\documentclass{article}

     \usepackage[nonatbib, preprint]{neurips_2020}

\usepackage[utf8]{inputenc} 
\usepackage[T1]{fontenc}    
\usepackage{hyperref}       
\usepackage{url}            
\usepackage{booktabs}       
\usepackage{amsfonts}       
\usepackage{nicefrac}       
\usepackage{microtype}      

\usepackage{graphicx}
\usepackage{multirow}
\usepackage{verbatim}

\graphicspath{ {./figures/} }
\usepackage{multirow}
\usepackage{makecell}
\usepackage{arydshln}
\usepackage{multicol}

\newcommand{\splitcell}[1]{\begin{tabular}{@{}l@{}}#1\end{tabular}}
\newcommand{\bsplitcell}[1]{$\left[\splitcell{#1}\right]$}

\usepackage{subfig}
\usepackage{wrapfig}
\usepackage{floatflt}
\usepackage{xcolor}

\usepackage[backend=biber,style=ieee, maxbibnames=99]{biblatex}
\title{Pyramidal Convolution: \\ Rethinking Convolutional Neural Networks for \\ Visual Recognition}

\author{%
  Ionut Cosmin Duta, Li Liu, Fan Zhu, Ling Shao \\
  Inception Institute of Artificial Intelligence (IIAI) \\
  \texttt{\{icduta\}@gmail.com}\\
}

\begin{document}

\maketitle

\begin{abstract}
This work introduces pyramidal convolution (PyConv), which is capable of processing the input at multiple filter scales. PyConv contains a pyramid of kernels, where each level involves different types of filters with varying size and depth, which are able to capture different levels of details in the scene. On top of these improved recognition capabilities, PyConv is also efficient and, with our formulation, it does not increase the  computational cost and  parameters compared to standard convolution. Moreover, it is very flexible and extensible, providing a large space of potential network architectures for different applications. PyConv has the potential to impact nearly every computer vision task and,  in this work, we present different architectures based on PyConv for four main tasks on visual recognition: image classification, video action classification/recognition, object detection and semantic image segmentation/parsing. Our approach shows  significant improvements over all these core tasks in comparison with the baselines. For instance, on image recognition, our 50-layers network outperforms in terms of recognition performance on ImageNet dataset its counterpart baseline ResNet with 152 layers, while having 2.39 times less parameters, 2.52 times lower computational complexity and more than 3 times less layers. On image segmentation, our novel framework sets a new state-of-the-art on the challenging ADE20K benchmark for scene parsing. 
\newline
 Code is available at: {\color{blue}\url{https://github.com/iduta/pyconv}}
\end{abstract}

\section{Introduction}
\begin{wrapfigure}{r}{0.4\textwidth}
  \centering
  \includegraphics[width=0.4\textwidth]{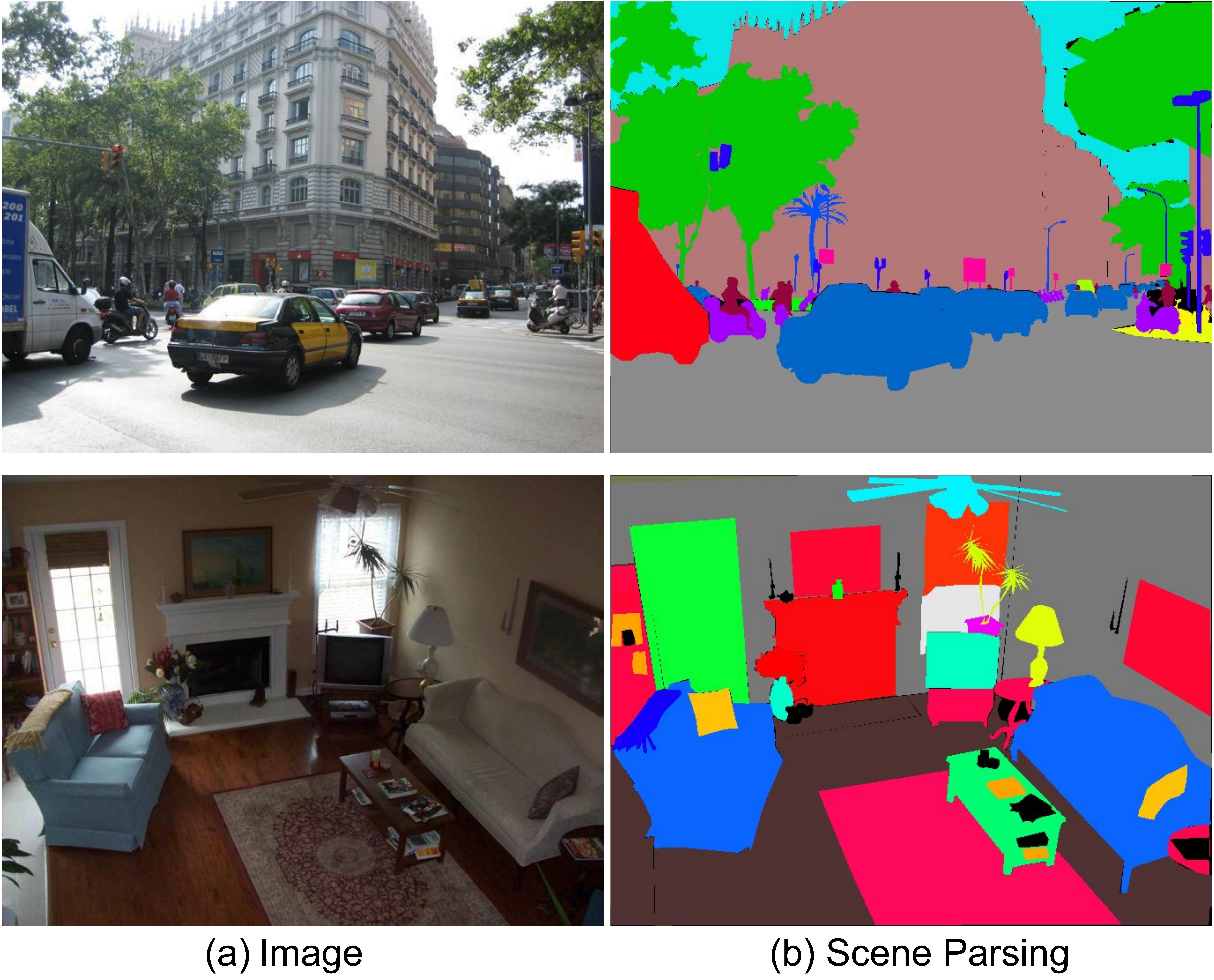}
  \caption{Scene parsing examples. An outdoor and indoor images with their associated pixel-level semantic category.}
  \label{fig:scene_parsing}
  \vspace{-0.2in}
\end{wrapfigure}

Convolutional Neural Networks (CNNs)~\cite{lecun1989backpropagation, lecun1998gradient} represent the workhorses of the most current computer vision applications. Nearly every recent state-of-the-art architecture for different tasks on visual recognition is based on CNNs~\cite{krizhevsky2012imagenet,simonyan2014very,szegedy2015going,ioffe2015batch,he2016deep,he2016identity,chollet2017xception,szegedy2017inception,huang2017densely,zoph2018learning,he2017mask,lin2017focal,lin2017feature,xie2017aggregated,hu2018squeeze,wu2018group}.  At the core of a CNN there is the convolution, which learns spatial kernels/filters for visual recognition. Most CNNs use a relatively small kernel size, usually  $3$$\times$$3$, forced by the fact that increasing the size comes with significant costs in terms of number of parameters and computational complexity. To cope with the fact that a small kernel size cannot cover a large region of the input, CNNs use a chain of  convolutions with small kernel size and downsampling layers, to gradually reduce the size of the input and to increase the receptive field of the network. However, there are two issues that can appear. First, even though for  many of current CNNs the theoretical receptive field can cover a big part of the input or even the whole input, in~\cite{zhou2014object} it is shown that the empirical receptive field is much smaller than the theoretical one, even more than 2.7 times smaller in the higher layers of the network. Second, downsampling the input without previously having access to enough context information (especially in complex scenes as in Fig.~\ref{fig:scene_parsing}) may affect the learning process and the recognition performance of the network, as useful details are lost since the receptive filed is not large enough to capture different dependencies in the scene before performing the downsampling.  
Natural images can contain extremely complicated scenes. Two examples (an outdoor and an indoor scenes in the wild) are presented in Fig.~\ref{fig:scene_parsing} on the left, while on the right we have the semantic label of each pixel in the scene (taken from  the ADE20K dataset~\cite{zhou2019semantic}). Parsing these images and providing a semantic category of each pixel is very challenging, however, it is one of the holy grail goals of the computer vision field. We can notice in these examples a big number of class categories in the same scene, some partially occluded, and different object scales.

We can see in  Fig.~\ref{fig:scene_parsing} that some class categories can have a very large spatial representations (e.g. buildings,  trees  or sofas) while other categories can have significantly smaller representations in the image (e.g. persons,  books or the bottle). Furthermore, the same object category can appear at different scales in the same image. For instance, the scales of cars in Fig.~\ref{fig:scene_parsing} vary significantly, from being one of the biggest objects in the image to cover just a very tiny portion of the scene. To be able to capture such a diversity of categories and such a variability in their scales, the use of a single type of kernel (as in standard convolution) with a single spatial size, may not be an optimal solution for such complexity.
One of the long standing goals of computer vision is the ability to process the input at multiple scales for capturing detailed information about the content of the scene.
One of the most notorious example in the hand-crafted features era is SIFT~\cite{lowe2004distinctive}, which extracts the descriptors at different scales. However, in the current deep learning era with learned features, the standard convolution is not implicitly equipped with the ability to process the input at multiple scales, and contains a single type of kernel with a single spatial size and depth. 

To address the aforementioned challenges, this work provides the following main contributions:
\textbf{(1)} We introduce pyramidal convolution (PyConv), which contains different levels of kernels with varying size and depth.
Besides enlarging the receptive field, PyConv can process the input using increasing kernel sizes in parallel, to capture different levels of details. On top of these advantages, PyConv is very efficient and, with our formulation, it can maintain a similar number of parameters and computational costs as the standard convolution. PyConv is very flexible and extendable, opening the door for a large variety of network architectures for numerous tasks of computer vision (see Section~\ref{section:pyconv}).
\textbf{(2)}~We propose two network architectures for image classification task that outperform the baselines by a significant margin. Moreover,  they are efficient in terms of number of parameters and computational costs and  can outperform other more complex architectures (see Section~\ref{sec:pyconvresnet}).
\textbf{(3)}~We propose a new framework for semantic segmentation. Our novel head for parsing the output provided by a backbone can capture different levels of context information from local to global. It provides state-of-the-art results on scene parsing (see Section~\ref{sec:pyconvseg}).
\textbf{(4)} We present network architectures based on PyConv for object detection and video classification tasks, where we report significant improvements in recognition performance over the baseline (see Appendix).

\section{Related Work}
Among the various methods employed for image recognition, the residual networks (ResNets) family~\cite{he2016deep,he2016identity,xie2017aggregated} represents one of the most influential and widely used. By using a shortcut connection, it facilitates the learning process of the network. These networks are used as backbones for various complex tasks, such as object detection and  instance segmentation \cite{he2016deep,he2017mask,lin2017focal,lin2017feature,xie2017aggregated,hu2018squeeze,wu2018group}. We use ResNets as baselines and make use of such architectures when building our different networks.

The seminal work  \cite{krizhevsky2012imagenet} uses a form of grouped convolution to distribute the computation of the convolution over two GPUs for overcoming the limitations of computational resources (especially memory). Furthermore, also \cite{xie2017aggregated} uses  grouped convolution but with the aim of improving the recognition performance in the ResNeXt architectures. We also make use of grouped convolution but in a different architecture.  
The works \cite{hu2018squeeze} and \cite{wang2018non} propose  squeeze-and-excitation and non-local blocks to capture context information. However, these are additional blocks that need to be inserted into the CNN; therefore, these approaches still need to use a spatial convolution in their overall CNN architecture (thus, they can be complementary to our approach). Furthermore, these blocks significantly increase the model and computational complexity.

On the challenging task of semantic segmentation, a very powerful network architecture is PSPNet~\cite{zhao2017pyramid}, which uses a pyramid pooling module (PPM)  head on top of a backbone in order to parse the scene for extracting different levels of details. Another powerful architecture is presented in \cite{chen2017rethinking}, which uses atrous spatial pyramid pooling (ASPP) head on top of a backbone. In contrast to  these competitive works, we propose a novel head for parsing the feature maps provided by a backbone, using a local multi-scale context aggregation module and a global multi-scale context aggregation block for efficient parsing of the input.  Our novel framework for image segmentation is not only very competitive in terms of recognition performance but is also significantly more efficient in terms of model and computational complexity than these strong architectures.

\section{Pyramidal Convolution \label{section:pyconv}}

\begin{wrapfigure}{r}{0.51\textwidth}
\vspace{-0.5in}
  \centering
  \includegraphics[width=0.51\textwidth]{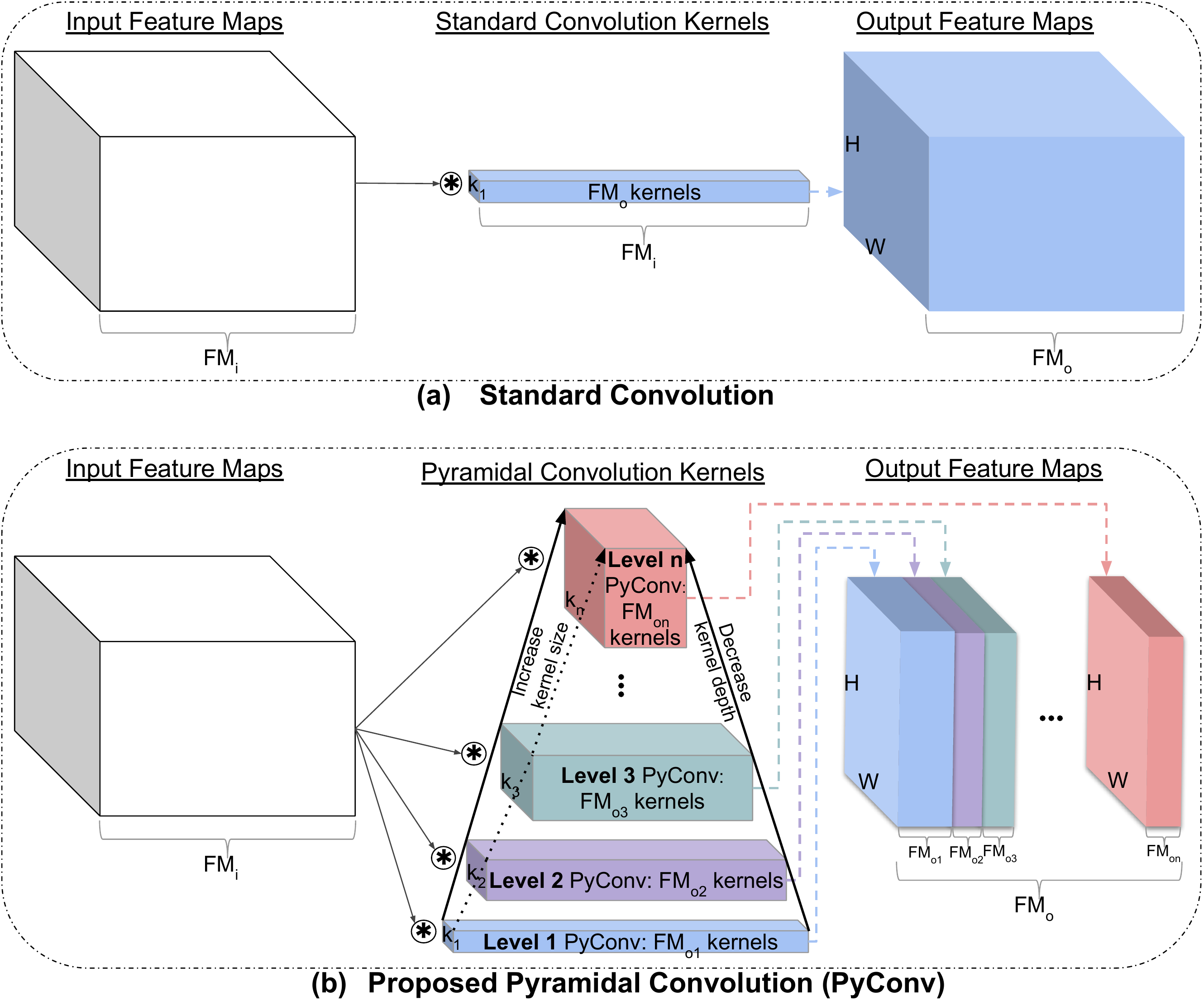}
  \caption{(a) Standard conv; (b) Proposed PyConv.}
  \label{fig:pyconv}
  \vspace{-0.2in}
\end{wrapfigure}

The standard convolution, illustrated in Fig.~\ref{fig:pyconv}(a), contains a single type of kernel: with a single spatial size ${K_1}^2$ (in the case of square kernels, e.g., height$\times$width: $3$$\times$$3$ = $3^2$, ${K_1}$ $=$ $3$) and the depth equal to the number of input feature maps $FM_{i}$. The result of applying a number of $FM_{o}$ kernels (all having the same spatial resolution and the same depth) over $FM_{i}$ input feature maps  is a number of $FM_{o}$ output feature maps (with spatial height $H$ and width $W$). Thus, the number of parameters and FLOPs (floating point operations) required for the standard convolution are: $parameters = {K_1}^2 \cdot FM_{i} \cdot FM_{o};$ $FLOPs = {K_1}^2 \cdot FM_{i} \cdot FM_{o} \cdot (W \cdot H).$

The proposed pyramidal convolution (PyConv), illustrated in Fig.~\ref{fig:pyconv}(b), contains a pyramid with $n$ levels of different types of kernels. The goal of the proposed PyConv is to process the input at different kernel scales without increasing the computational cost or the model complexity (in terms of parameters). At each level of the PyConv, the kernel contains a different spatial size, increasing kernel size from the bottom of the pyramid (level 1 of PyConv) to the top (level $n$ of PyConv). Simultaneously with increasing the spatial size, the depth of the kernel is decreased from level 1 to level $n$. Therefore, as shown in Fig.~\ref{fig:pyconv}(b), this results in two interconnected pyramids, facing opposite directions. One pyramid has the base at the bottom (evolving to the top by decreasing the kernel depth) and the other pyramid has the base on top, where the convolution kernel has the largest spatial size (evolving to the bottom by decreasing the spatial size of the kernel).
\begin{wrapfigure}{r}{0.55\textwidth}
\vspace{-0.1in}
  \centering
  \includegraphics[width=0.55\textwidth]{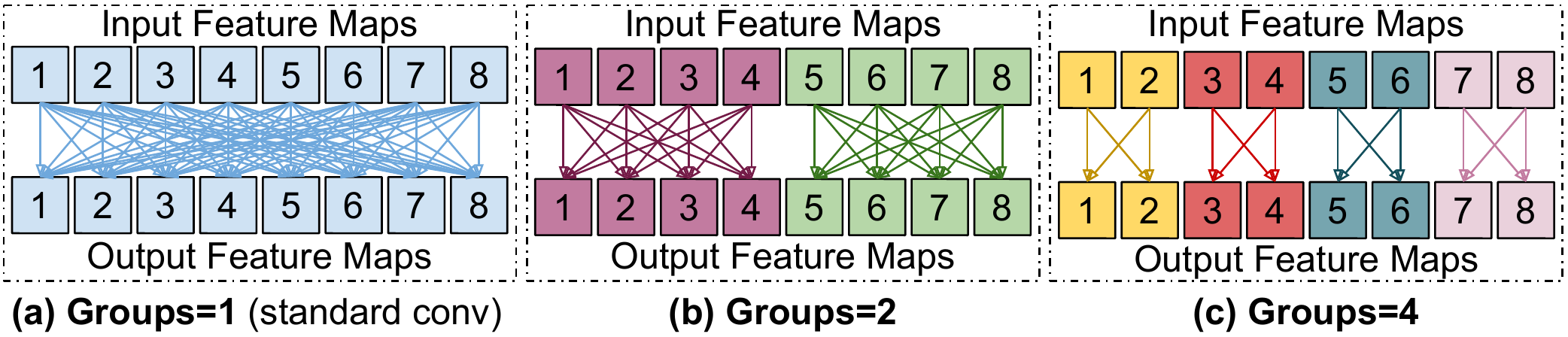}
  \caption{Grouped Convolution.}
  \label{fig:grouped_conv}
\end{wrapfigure}

\vspace{-0.15in}
To be able to use different depths of the kernels at each level of PyConv, the input feature maps  are split into different groups, and apply the kernels independently for each input feature maps group. This is called grouped convolution,  an illustration is presented in Fig.~\ref{fig:grouped_conv}  where we show three examples (the color encodes the group assignment). In these examples, there are eight input and output feature maps. Fig.~\ref{fig:grouped_conv}(a) shows the case comprising a single group of input feature maps, this is the standard convolution, where the depth of the kernels is equal to the number of input feature maps.  In this case, each output feature map is connected to all input feature maps.  Fig.~\ref{fig:grouped_conv}(b) shows the case when the input feature maps are split into two groups, where the kernels are applied independently over each group, therefore, the depth of the kernels is reduced by two. As shown in  Fig.~\ref{fig:grouped_conv},  when the number of groups is increased,  the connectivity (and thus the depth of the kernels) decreases. As a result, the number of parameters and the computational cost of a convolution is reduced by a factor equal to the number of groups.


As illustrated in Fig.~\ref{fig:pyconv}(b), for the input feature maps $FM_{i}$, each level of the PyConv $\{1, 2, 3, ..., n\}$ applies different kernels with a different spatial size for each level $\{{K_1}^2$, ${K_2}^2$, ${K_3}^2$, ...,  ${K_n}^2\}$ and  with different kernel depths  \scalebox{0.8}{$\{FM_{i}$, $\frac{FM_{i}}{( \frac{{K_2}^2}{{K_1}^2})}$, $\frac{FM_{i}}{( \frac{{K_3}^2}{{K_1}^2})}$, ...,  $\frac{FM_{i}}{( \frac{{K_n}^2}{{K_1}^2})}\}$}, which outputs a different number of output feature maps \scalebox{0.8}{$\{FM_{o_1}$,  $FM_{o_2}$, $FM_{o_3}$, ..., $FM_{o_n}\}$}  (with height $H$ and width $W$).  Therefore, the number of parameters and the computational cost (in terms of FLOPs) for PyConv are:
\renewcommand{\arraystretch}{0.5}
\noindent\begin{minipage}{.4\textwidth}
\begin{equation}
\scalebox{0.77}{$%
\begin{array}{|l||l|}
parameters = & FLOPs = \\
\vspace{-5pt}
{K_n}^2 \cdot \frac{FM_{i}}{( \frac{{K_n}^2}{{K_1}^2})} \cdot FM_{o_n}+ 
& {K_n}^2 \cdot \frac{FM_{i}}{( \frac{{K_n}^2}{{K_1}^2})} \cdot FM_{o_n} \cdot (W \cdot H)+ \vspace{-5pt}\\
\vspace{-4pt}
 \vdots & \vdots \vspace{+5pt} \\
\vspace{+3pt}
{K_3}^2 \cdot \frac{FM_{i}}{( \frac{{K_3}^2}{{K_1}^2})} \cdot FM_{o_3} + 
& {K_3}^2 \cdot \frac{FM_{i}}{( \frac{{K_3}^2}{{K_1}^2})} \cdot FM_{o_3} \cdot (W \cdot H)+ \\

{K_2}^2 \cdot \frac{FM_{i}}{( \frac{{K_2}^2}{{K_1}^2})} \cdot FM_{o_2} + 
& {K_2}^2 \cdot \frac{FM_{i}}{( \frac{{K_2}^2}{{K_1}^2})} \cdot FM_{o_2} \cdot (W \cdot H)+ \\

{K_1}^2 \cdot FM_{i} \cdot FM_{o_1};  
&{K_1}^2 \cdot FM_{i} \cdot FM_{o_1} \cdot (W \cdot H), \\ 

\end{array}
 \label{eq:pyconv}
 $
 }
\end{equation}
\end{minipage}
$\;\;\;\;\;\;\;\;\;\;\;\;$
\noindent\begin{minipage}{.5\textwidth}
where \scalebox{0.8}{$FM_{o_n} + \cdot \cdot \cdot + FM_{o_3} + FM_{o_2} + FM_{o_1} =  FM_{o}$}. Each row in these equations represents the number of parameters and computational cost for a level in PyConv. 
If each level of PyConv outputs an equal number of feature maps, then the number of parameters and the computational cost of PyConv are distributed evenly along each pyramid level.
\end{minipage}
\vspace{-1pt}
With this formulation, regardless of the number of levels of PyConv and the continuously increasing kernel spatial sizes from ${K_1}^2$ to ${K_n}^2$,  the computational cost and the number of parameters is similar to the standard convolution with a single  kernel size ${K_1}^2$.
To link the illustration in  Fig.~\ref{fig:grouped_conv} with Equations~\ref{eq:pyconv}, the denominator of $ FM_{i} $ in Equations~\ref{eq:pyconv} refers to the number of groups (G) that the input feature maps $FM_{i}$ are split in Fig.~\ref{fig:grouped_conv}.

In practice, when building a PyConv there are several additional rules. The denominator of $FM_{i}$ at each level of the pyramid in Equations~\ref{eq:pyconv}, should be a divisor of $FM_{i}$. In other words, at each pyramid level, the number of feature maps from each created group should be equal. Therefore, as an approximation, when choosing the number of groups for each level of the pyramid (and thus the depth of the kernel), we can take the closest number to the denominator of $FM_{i}$ from the list of possible divisors of $FM_{i}$. Furthermore, the number of groups for each level should be also a divisor for the number of output feature maps of each level of PyConv. To be able to easily create  different network architectures with PyConv, it is recommended that the number of input feature maps, the groups for each level of pyramid, and the number of output feature maps for each level of PyConv, to be numbers of power of 2. Next sections show practical examples.

The main advantages of the proposed PyConv are:
(1)~{\bf Multi-Scale Processing.} Besides the fact that, compared to the standard convolution,  PyConv can enlarge the receptive field of the kernel without additional costs,  it also applies in parallel different types of kernels, having different spatial resolutions and depths. Therefore, PyConv parses the input at multiple scales capturing more detailed information. This  double-oriented  pyramid of kernels types, where on one side the kernel sizes are increasing and on the other side the kernel depths (connectivity) are decreasing (and vice-versa), allows PyConv to  provide a very diverse pool of combinations of different kernel types that the network can explore during learning. The network can explore from large receptive fields of the kernels with lower connectivity to smaller receptive fields with higher connectivity.
These different types of kernels of PyConv bring complementary information and help boosting the recognition performance of the network. The kernels with smaller receptive field can focus on details, capturing information about smaller objects and/or parts of the objects, while increasing the kernels size provides more reliable details about larger objects and/or context information.

(2) {\bf Efficiency.} In comparison with the standard convolution, PyConv maintains, by default, a similar number of model parameters and requirements in computational resources, as shown in Equation~\ref{eq:pyconv}. Furthermore, PyConv offers a high degree of parallelism due to the fact that the pyramid levels can be  independently  computed in parallel. Thus, PyConv can also offer the possibility of customizable heavy network architectures (in the case where the architecture cannot fit into the memory of a computing unit and/or it is too expensive in terms of FLOPs), where the levels of PyConv can be executed independently on different computing units and then the outputs can be merged.

(3) {\bf Flexibility.} PyConv opens the door for a great variety of network architectures. The user has the flexibility to choose the number of layers of the pyramid, the kernel sizes and depths at each PyConv level, without paying the price of increasing the number of parameters or the computational costs. Furthermore, the number of output feature maps can be different at each level.  For instance, for a particular final task it may be more useful to have less output feature maps from the kernels with small receptive fields and more output feature maps from the kernels with bigger receptive fields. Also, the PyConv settings can be different along the network, thus, at each layer of the network we can have different PyConv settings. For example, we can start with several layers for PyConv, and  based on the resolution of the input feature maps at each layer of the network, we can decrease the levels of PyConv as the resolution decreases along the network. That being said, we can now build architectures using PyConv for different visual recognition tasks.

\section{PyConv Networks for Image Classification \label{sec:pyconvresnet}}

\begin{wrapfigure}{r}{0.13\textwidth}
\vspace{-0.4in}
  \centering
    \includegraphics[width=0.13\textwidth]{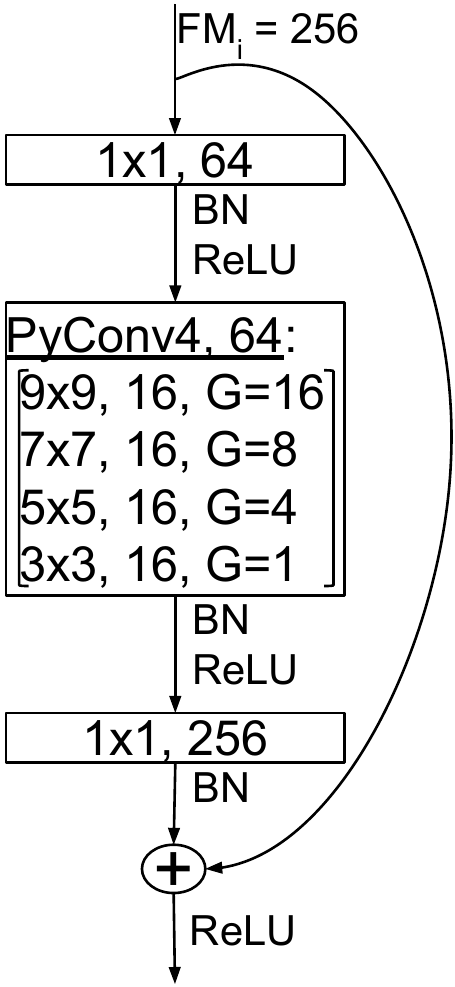}
  \caption{PyConv bottleneck building block.\label{fig:pyconv_block}}
  \vspace{-0.2in}
\end{wrapfigure}

For our PyConv network architectures on image classification, we use a residual bottleneck building block similar to the one reported  in~\cite{he2016deep}. Fig.~\ref{fig:pyconv_block} shows an example of a building block used on the first stage of our network. First, it applies a $1$$\times$$1$ conv to reduce the input feature maps to 64, then we use our proposed PyConv with four levels of different kernels sizes: $9$$\times$$9$, $7$$\times$$7$, $5$$\times$$5$, $3$$\times$$3$. Also the depth of the kernels varies along each level, from 16 groups to full depth/connectivity. Each level outputs 16 feature maps, which sums a 64  output feature maps for the PyConv. Then a $1$$\times$$1$ conv is applied to regain the initial number of feature maps. As common, batch normalization~\cite{ioffe2015batch} and ReLU activation function~\cite{nair2010rectified} follow a conv block. Finally, there is a shortcut connection that can help with the identity mapping.

Our proposed network for image classification, {\bf PyConvResNet}, is illustrated in Table~\ref{table:net}. For direct comparison we place aside also the baseline architecture ResNet~\cite{he2016deep}.  Table~\ref{table:net} presents the case for a 50-layers deep network, for the other depths, the number of layers are increased as in~\cite{he2016deep}. Along the network we can identify four main stages, based on the spatial size of the feature maps. For PyConvResNet, we start with a PyConv with four levels. Since the spatial size of the feature maps decreases at each stage, we reduce also the PyConv levels. On the last main stage, the network ends up with only one level for PyConv, which is basically the standard convolution. This is appropriate because the spatial size of the feature maps is only $7$$\times$$7$, thus, three successive convolutions  of  size $3$$\times$$3$ cover well the feature maps.  Regarding the efficiency, PyConvResNet provides also a slight decrease in FLOPs.

\renewcommand{\arraystretch}{0.9}
\addtolength{\tabcolsep}{-5.5pt}
\begin{wraptable}{r}{0.5\textwidth}
\vspace{-0.2in}
\centering
\caption{\small PyConvResNet and PyConvHGResNet.}
\scalebox{0.58}{
\begin{tabular}{c|c|c|c|c}
\hline
stage                  & 
output                 & 
ResNet-50             & 
PyConvResNet-50     & 
PyConvHGResNet-50 \\ 
\hline

\multirow{2}{*}{}                 &
112$\times$112                & 
7$\times$7, 64, s=2     & 
7$\times$7, 64, s=2      & 
7$\times$7, 64, s=2    \\
\cline{2-5}
 & 
 & 
3$\times$3 max pool,s=2&  
 &  \\  
 \hline
 
1&
56$\times$56 & 
\bsplitcell{1$\times$1, 64\\ 3$\times$3, 64\\ 1$\times$1, 256}$\times$3 &
\bsplitcell{1$\times$1, 64\\ \underline{PyConv4, 64:}\\  \bsplitcell{9$\times$9, 16, G=16\\ 7$\times$7, 16, G=8 \\ 5$\times$5, 16, G=4 \\ 3$\times$3, 16, G=1} \\ 1$\times$1, 256}$\times$3 &
\bsplitcell{1$\times$1, 128\\ \underline{PyConv4, 128:}\\  \bsplitcell{9$\times$9, 32, G=32\\ 7$\times$7, 32, G=32 \\ 5$\times$5, 32, G=32 \\ 3$\times$3, 32, G=32} \\ 1$\times$1, 256}$\times$3 \\ 
\hline

2 &
28$\times$28 &
\bsplitcell{1$\times$1, 128\\ 3$\times$3, 128\\ 1$\times$1, 512}$\times$4 &
\bsplitcell{1$\times$1, 128\\ \underline{PyConv3, 128:}\\  \bsplitcell{7$\times$7, 64, G=8 \\ 5$\times$5, 32, G=4 \\ 3$\times$3, 32, G=1}\\ 1$\times$1, 512}$\times$4 &
\bsplitcell{1$\times$1, 256\\ \underline{PyConv3, 256:}\\  \bsplitcell{7$\times$7, 128, G=64 \\ 5$\times$5, 64, G=64 \\ 3$\times$3, 64, G=32}\\ 1$\times$1, 512}$\times$4 \\ 
\hline

3 &
14$\times$14 &
\bsplitcell{1$\times$1, 256\\ 3$\times$3, 256\\ 1$\times$1, 1024}$\times$6 &
\bsplitcell{1$\times$1, 256\\ \underline{PyConv2, 256:}\\  \bsplitcell{5$\times$5, 128, G=4 \\ 3$\times$3, 128, G=1}\\ 1$\times$1, 1024}$\times$6 &
\bsplitcell{1$\times$1, 512\\ \underline{PyConv2, 512:}\\  \bsplitcell{5$\times$5, 256, G=64 \\ 3$\times$3, 256, G=32}\\ 1$\times$1, 1024}$\times$6 \\ \hline

4 &
7$\times$7 &
\bsplitcell{1$\times$1, 512\\ 3$\times$3, 512\\ 1$\times$1, 2048}$\times$3 &
\bsplitcell{1$\times$1, 512\\ \underline{PyConv1, 512:}\\  \bsplitcell{3$\times$3, 512, G=1}\\ 1$\times$1, 2048}$\times$3 & 
\bsplitcell{1$\times$1, 1024\\ \underline{PyConv1, 1024:}\\  \bsplitcell{3$\times$3, 1024, G=32}\\ 1$\times$1, 2048}$\times$3 \\ 
\hline

 &
1$\times$1 &
\begin{tabular}[c]{@{}c@{}}global avg pool\\ 1000-d fc\end{tabular} &
\begin{tabular}[c]{@{}c@{}}global avg pool\\ 1000-d fc\end{tabular} &
\begin{tabular}[c]{@{}c@{}}global avg pool\\ 1000-d fc\end{tabular} \\ 
\hline

\multicolumn{2}{c|}{\# params}                  &
\bf{25.56}  $\times$ $10^6$                 &
\bf{24.85}  $\times$ $10^6$                    &
\bf{25.23}  $\times$ $10^6$                   \\
\hline

\multicolumn{2}{c|}{FLOPs}                      & 
\bf{4.14}  $\times$ $10^9$                   &
\bf{3.88}  $\times$ $10^9$&
\bf{4.61}  $\times$ $10^9$             \\
\hline

\end{tabular}}
\label{table:net}
\vspace{-0.22in}
\end{wraptable}
\addtolength{\tabcolsep}{5.5pt}
\renewcommand{\arraystretch}{0.7}

As we highlighted, flexibility is a strong point of PyConv, Table~\ref{table:net} presents another architecture based on PyConv, {\bf PyConvHGResNet}, which uses a higher grouping for each level. For this architecture we set a minimum of 32 groups and a maximum of 64 in the PyConv. The number of feature maps for the spatial convolutions is doubled to provide better capabilities on learning spatial filters.  Note that for  stage one of the network, it is not possible to increase the number of groups more than 32 since this is the number of input and output feature maps for each level. Thus, PyConvHGResNet produces a slight increase in FLOPs.

As our networks contain different levels of kernels, it can perform the downsampling of the feature maps using different kernel sizes. This is important as downsampling produces loss of spatial resolution and therefore loss of details, but having different kernel sizes to perform the downsampling can take into account different levels of spatial context dependencies to perform the dowsampling in parallel. As can be seen in Table~\ref{table:net}, the original ResNet~\cite{he2016deep} uses a max pooling layer before the first stage of the network to downsample the feature maps and to get the translation invariance. Different from the original ResNet, we move the max pooling on the first projection shortcut,  just before the $1$$\times$$1$ conv (usually, the first shortcut of a stage contains a projection $1$$\times$$1$ conv to adapt the number of feature maps and their spatial resolution for the summation with the output of the block). This is similar to projection shortcut in~\cite{duta2020improved}.
Therefore, for the original ResNet, the downsampling is not performed by the first stage (as the max pooling performs this before), the next three main stages perform the downsampling on their first block. In our networks, all four main stages perform the downsampling in their first block. 

This change does not increase the number of parameters of the network and does not affect significantly the computational costs (as can be seen in Table~\ref{table:net}, as the first block uses the spatial convolutions with the stride 2), providing advantages in recognition performance for our networks.
Moving the max pooling to the shortcut gives our approach the opportunity  to have access to larger spatial resolution of the feature maps in the first block of the first stage, to downsample the input using multiple kernel scales and, at the same time, to benefit from the translation invariance provided by max pooling. The results show that our networks provide improved recognition capabilities.

\section{PyConv Network on Semantic Segmentation\label{sec:pyconvseg}}

\renewcommand{\arraystretch}{1}
\addtolength{\tabcolsep}{-2pt}
\begin{figure}[t]
\parbox{0.74\textwidth}{
  \centering
  \includegraphics[width=0.78\textwidth]{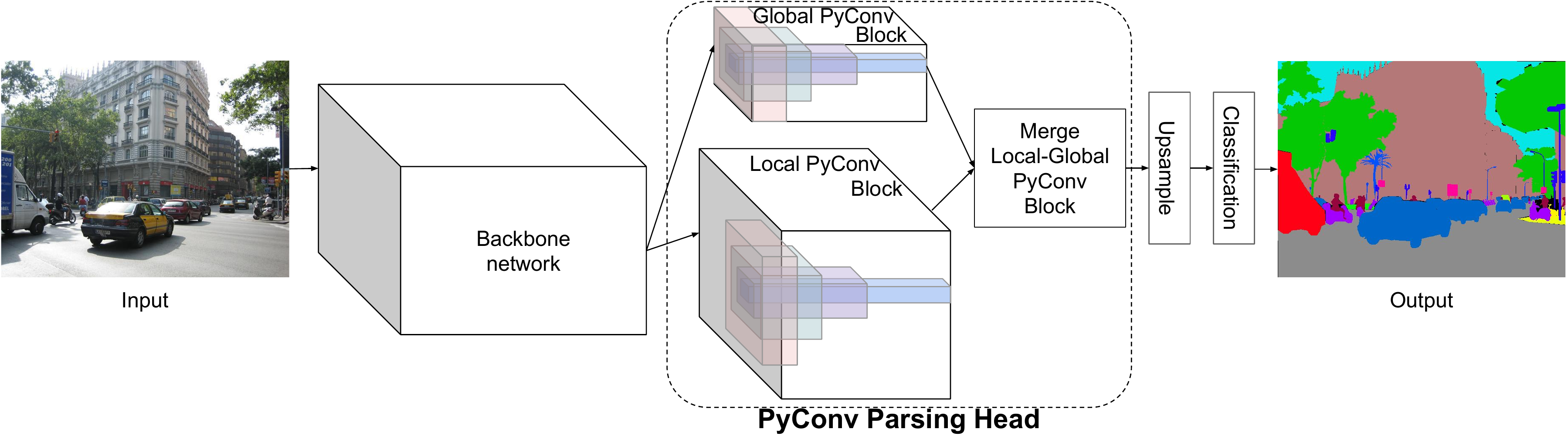}
  \caption{PyConvSegNet framework for image segmentation. \label{fig:pyconvseg}}}$\;$
\parbox{0.28\textwidth}{
  \centering
  \includegraphics[width=0.2\textwidth]{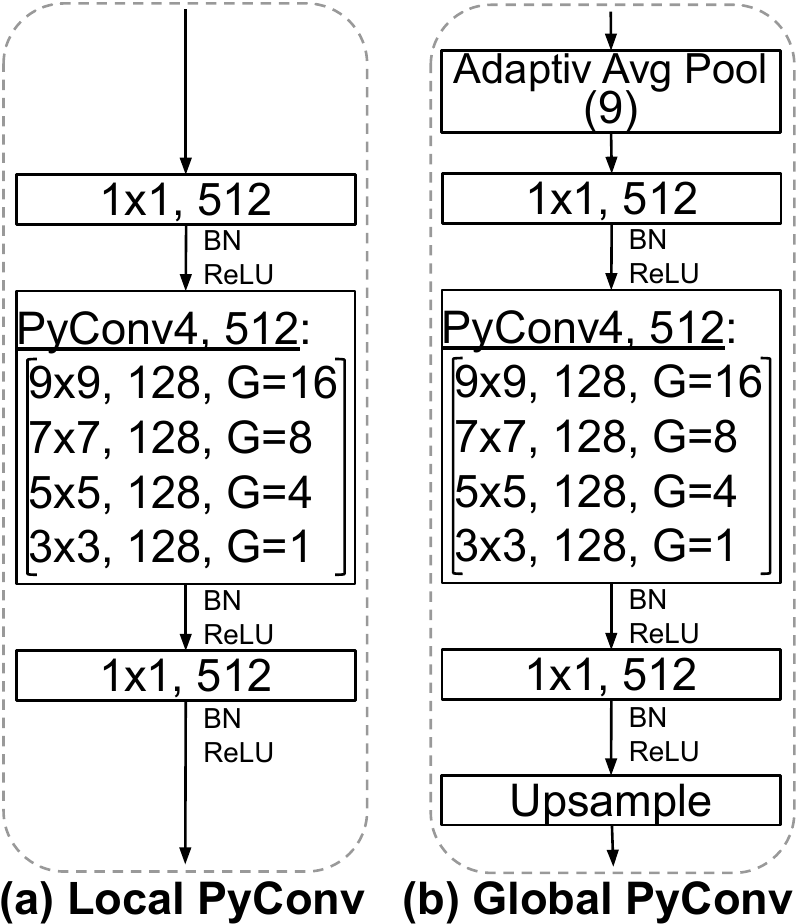}
  \caption{PyConv blocks. \label{fig:parse_pyconv}}}
\end{figure}
\addtolength{\tabcolsep}{2pt}
\renewcommand{\arraystretch}{1}

Our proposed framework for scene parsing (image segmentation) is illustrated in Fig.~\ref{fig:pyconvseg}.  To build an effective pipeline for scene parsing, it is necessary to create a head that can parse the feature maps provided by the backbone and obtain not only local but also global information.  The head should be able to deal with fine details and, at the same time, take into account the context information. We propose a novel head for scene parsing (image segmentation) task, PyConv Parsing Head (PyConvPH). The proposed  PyConvPH is able to deal with both local and global information at multiple scales.

PyConvPH contains three main components: (1) {\bf Local PyConv block} (LocalPyConv), which is mostly responsible for smaller objects and  capturing local fine details at multiple scales, as shown in Fig.~\ref{fig:pyconvseg}. It also applies  different type of kernels, with different spatial sizes and depths, which can also be seen as a local multi-scale context aggregation module. The detailed information about each component of LocalPyConv is represented in Fig.~\ref{fig:parse_pyconv}(a). LocalPyConv takes the output feature maps from the backbone and applies a 1$\times$1 conv to reduce the number of feature maps to 512. Then, it performs a PyConv with four layers to capture different local details at four scales of the kernel  9$\times$9,  7$\times$7, 5$\times$5, and 3$\times$3. Additionally,  the kernels have different connectivity, represented by the number of groups (G). Finally, it applies a 1$\times$1 conv to combine the information extracted at different kernel sizes and depths. As is standard, all convolution blocks are followed by a batch normalization layer \cite{ioffe2015batch} and a ReLU activation function~\cite{nair2010rectified}.

(2) {\bf Global PyConv block} (GlobalPyConv) is responsible for capturing global details about the scene, and for dealing with very large objects. It is a multi-scale global aggregation module. The components of GlobalPyConv are represented in Fig.~\ref{fig:parse_pyconv}(b). As the input image size can vary, to ensure that we can capture full global information we keep the largest spatial size dimension as 9.
We apply an adaptive average pooling that reduces the spatial size of the feature maps to 9$\times$9 (in the case of square images), which  still maintains reasonable spatial resolution. Then we apply a 1$\times$1 conv to reduce the number of feature maps to 512. We use a PyConv with four layers similarly as in the LocalPyConv. However, as now we have decreased the spatial size of the feature maps to 9$\times$9, the PyConv kernels can cover very large parts of the input, ultimately, the  layer with a 9$\times$9 convolution covers the whole input and captures full global information, as illustrated also in Fig.~\ref{fig:pyconvseg}. Then we apply a 1$\times$1 conv to fuse the information from different scales. Finally, we upsample the feature maps to the initial size before the adaptive average pooling, using a  bilinear~interpolation.

(3) {\bf Merge Local-Global PyConv block} performs first the concatenation of the output feature maps from the LocalPyConv and GlobalPyConv blocks. Over the resulting 1024 feature maps, it applies a PyConv with one level, which is basically a standard 3$\times$3 conv that outputs 256 feature maps. We use here a single level for PyConv because the previous layers already captured all levels of context information and it is more important at this point to focus on merging this information (thus, to use full connectivity of the kernels) as it approaches the final classification stage. To provide the final output, the framework continues with an upsample layer (using also bilinear interpolation) to restore the feature maps to the initial input image size; finally, there is a classification layer which contains a 1$\times$1 conv, to provide the output with a dimension equal to the number of classes. As illustrated  in Fig.~\ref{fig:pyconvseg}, our proposed framework is able to capture local and global information at multiple scales of kernels, parsing the image and providing a strong representation. Furthermore, our framework is also very efficient, and in the following we provide the exact numbers and the comparison with other state-of-the-art frameworks.

\section{Experiments \label{sec:exp}}
{\bf Experimental setup.}  For image classification task we perform our experiments on the commonly used ImageNet  dataset~\cite{russakovsky2015imagenet}. It consists of 1000 classes, 1.28 million training images and 50K  validation images. We report both top-1 and top-5 error rates. We follow the settings in \cite{goyal2017accurate,he2016deep,he2016identity} and use the SGD optimizer with a standard momentum of 0.9, and weight decay of 0.0001.  We train the model for 90 epochs, starting with a learning rate of 0.1 and reducing it by 1/10 at the 30-th, 60-th and 80-th epochs, similarly to \cite{he2016deep,goyal2017accurate}. The models are trained using 8 GPUs V100. We use the standard 256 training mini-batch size and data augmentation as in~\cite{szegedy2015going,goyal2017accurate}, training/testing on $224$$\times$$224$ image crop.
For image segmentation we use ADE20K benchmark \cite{zhou2019semantic}, which is one of the most challenging datasets for image segmentation/parsing. It contains 150 classes and a high level of scenes diversity, containing both object and stuff classes. It is divided in 20K/2K/3K images for training, validation and testing.
As standard, we report both pixel-wise accuracy (pAcc.) and mean of class-wise intersection over union (mIoU). We train for 100 epochs with a mini-batch size of 16 over 8 GPUs V100, using train/test image crop size of $473$$\times$$473$. We follow the training settings as  in~\cite{zhao2017pyramid}, including the auxiliary loss, with the weight 0.4. 


\addtolength{\tabcolsep}{-5pt}
\renewcommand{\arraystretch}{1.}
\begin{wraptable}{r}{0.4\textwidth}
\vspace{-0.2in}
\centering
\caption{ImageNet ablation experiments of PyConvResNet.}
\label{table:ablation_exp}
\scalebox{0.8}{
\begin{tabular}{l|cccc}
\hline
PyConv levels&top-1{\small (\%)}&top-5{\small(\%)}& \small params &\tiny GFLOPs\\
\hline
(1, 1, 1, 1){\small baseline}&23.88&7.06&25.56&4.14 \\
(2, 2, 2, 1)& 23.12 & 6.58 &24.91 &3.91 \\
(3, 3, 2, 1)& 22.98 &6.62 &24.85 &3.85 \\ 
(4, 3, 2, 1)&22.97 & 6.56&24.85&3.84 \\ 
top(4, 3, 2, 1)&23.18 & 6.60&24.24&3.63 \\ 
(5, 4, 3, 2) & 23.03 & 6.56&23.45&3.71\\
\hline
(4, 3, 2, 1) max &22.46 & 6.24&24.85&3.88 \\
(4, 3, 2, 1) final&22.12 & 6.20&24.85&3.88
\end{tabular}}
\vspace{-0.2in}
\end{wraptable}
\addtolength{\tabcolsep}{5pt}

\textbf{PyConv results on image recognition}. We present in  Table~\ref{table:ablation_exp}  the ablation experiments results of the proposed PyConv for image recognition task on the ImageNet dataset where, using the network with 50 layers, we vary the number of levels of PyConv. We first provide a direct comparison  to the baseline ResNet~\cite{he2016deep} without any additional changes, just replacing the standard 3x3 conv with our PyConv.   The column "PyConv levels" points to the number of levels used at each of the four main stages of the network. The PyConv levels (1, 1, 1, 1) represent the case when we use a single level for PyConv on all four stages, which is basically the baseline ResNet. Remarkably, just increasing the number of PyConv levels to two  provides a significant improvement in recognition performance, improving the top-1 error rate from 23.88 to 23.12. At the same time it requires less number of parameters and FLOPS than the baseline. Note that by just using two levels for PyConv ($5$$\times$$5$ and $3$$\times$$3$ kernels), it has already significantly increased the receptive field at each stage of the network. Gradually increasing the levels of PyConv at each level brings further improvement, for the PyConv levels (4, 3, 2, 1) it brings the top-1 error rate to 22.97 with even lower number of parameters and FLOPs. We also run the experiment using only the top level of the PyConv at each main stage network, basically the opposite case of the baseline which uses only the bottom level. Therefore, top(4, 3, 2, 1) refers to the case when using only the fourth level of the PyConv for the stage 1 ($9$$\times$$9$ kernel), third level for stage 2 ($7$$\times$$7$ kernel), second level for stage 3  ($5$$\times$$5$ kernel) and first level of stage 4 ($3$$\times$$3$ kernel). This configuration also provides significant improvements in recognition performance compared to the baseline while having a lower number of parameters and FLOPs, showing that our formulation of increasing the kernel sizes for building the network is beneficial in many possible configurations.

We also add one more layer to PyConv for each stage of the network, (5, 4, 3, 2) case, where the fifth level has a $11$$\times$$11$ kernel, but we do not notice further improvements in recognition performance. In the rest of the paper we use (4, 3, 2, 1) levels of PyConv for image classification task. However, we find  this configuration reasonably good for this task with the input image resolution (224$\times$224), however, if, for instance, the input image resolution is increased, then other settings of PyConv may provide even further improvements.  Moving the max pooling to the shortcut, which provides access for PyConv to perform the downsampling at multiple kernel sizes, improves further the  top-1 error rate to 22.46. To further benefit from the translation invariance and to address the fact that a $1$$\times$$1$ conv lacks the spatial resolution for performing downsampling, we maintain a max pooling on the projection shortcut in the first block of each following stages. Our final network result is 22.12 top-1 error requiring only 24.85 million parameters and 3.88 {\small GFLOPs}. 
The conclusion is that regardless of the settings of PyConv, using our formulation, it consistently provides better results than the baseline.

\begin{figure*}[t]
\centering
\begin{tabular}{ccc}
\subfloat{\includegraphics[width=0.3\textwidth]{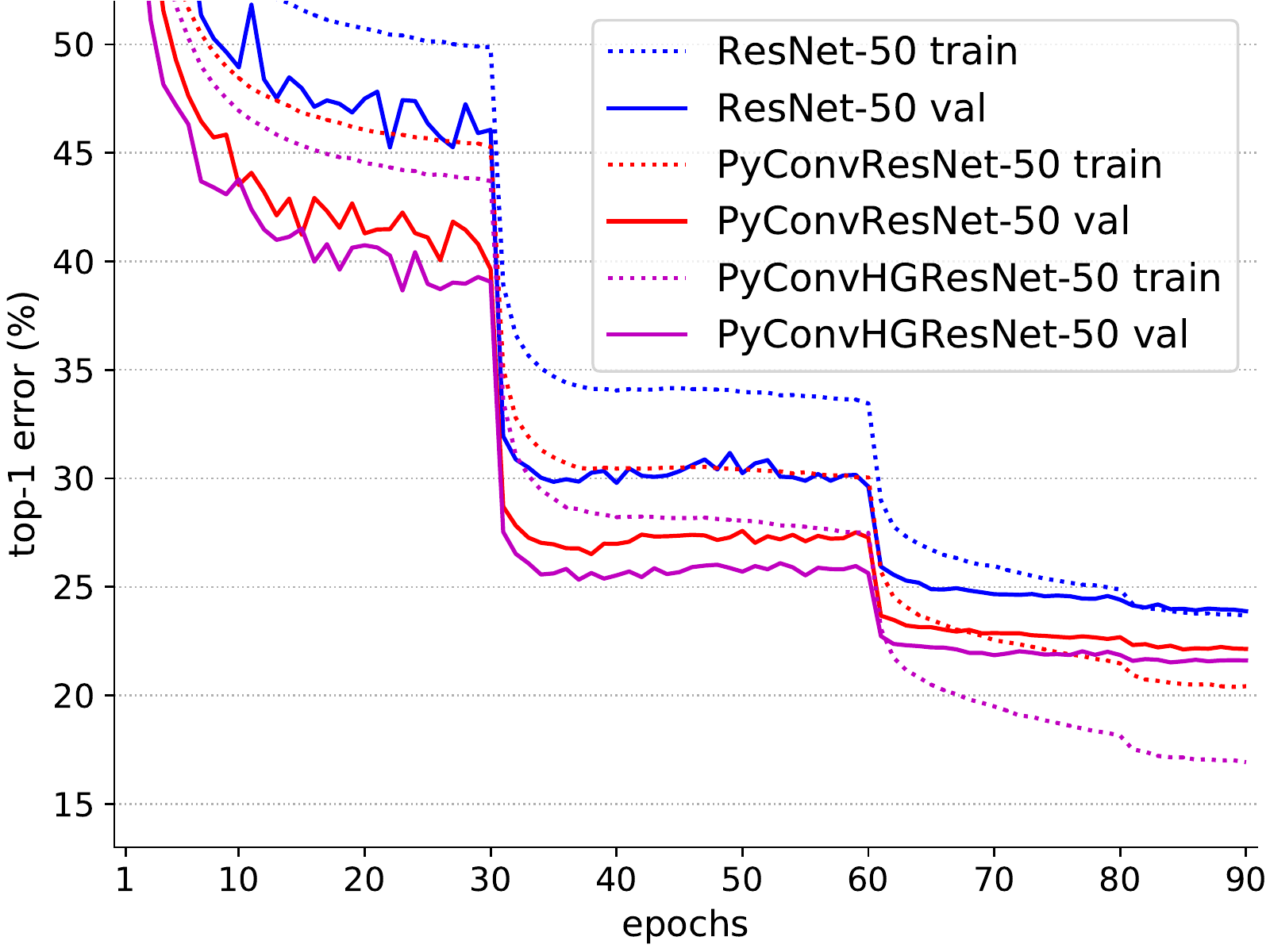}} &
\subfloat{\includegraphics[width=0.3\textwidth]{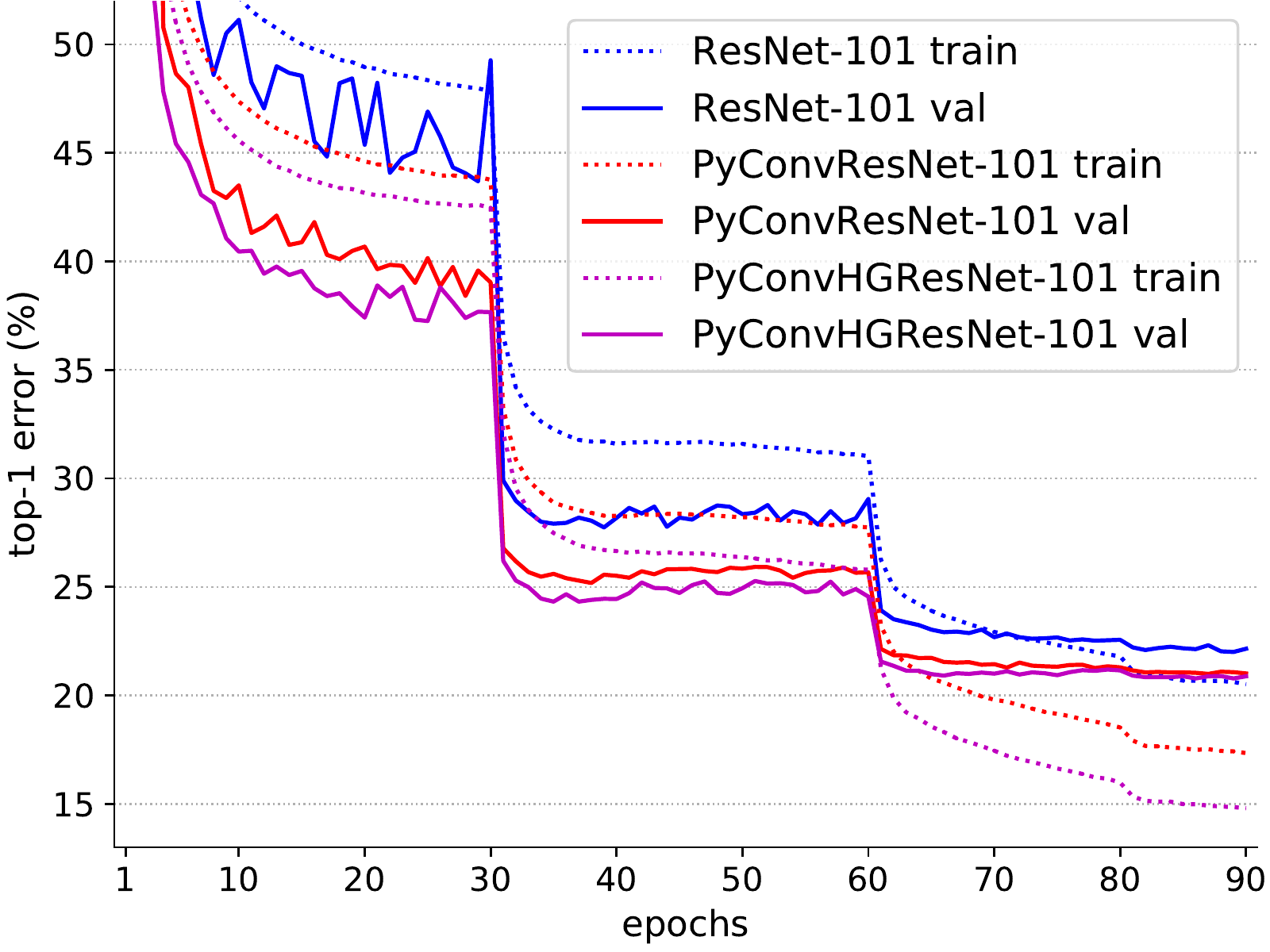}} &
\subfloat{\includegraphics[width=0.3\textwidth]{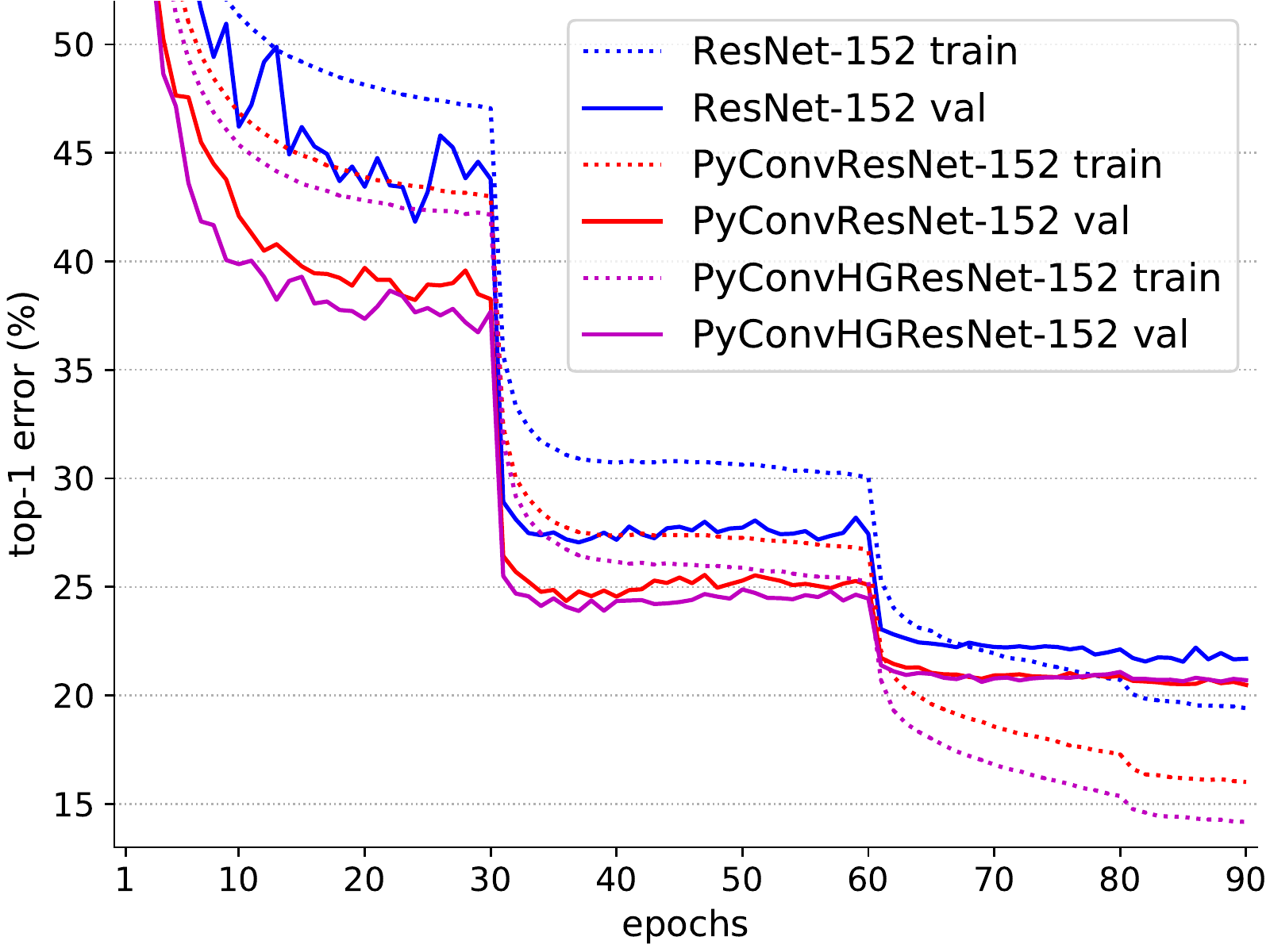}}
\end{tabular}
\caption{ImageNet training curves for ResNet and PyConvResNet on 50, 101 and 152 layers.}
\label{fig:curves_pyconv}
\end{figure*}

\addtolength{\tabcolsep}{-1.9pt}
\begin{table}[t]
\vspace{-0.1in}
\centering
\caption{Validation error rates comparison results of PyConv on ImageNet with other architectures.}
\label{table:pyconv_sota}
\scalebox{0.85}{
\begin{tabular}{l|cccc|cccc|cccc}
\hline
\multirow{2}{*}{Network}&\multicolumn{4}{c|}{network depth: 50 }&\multicolumn{4}{c|}{network depth: 101 }&\multicolumn{4}{c}{network depth: 152}\\ \cline{2-13}
&top-1&top-5&\scalebox{0.8}{params}&\scalebox{0.7}{GFLOPs}&top-1&top-5&\scalebox{0.8}{params} &\scalebox{0.7}{GFLOPs}&top-1&top-5&\scalebox{0.8}{params} &\scalebox{0.7}{GFLOPs}\\ \hline
ResNet (baseline)\cite{he2016deep}&23.88&7.06&25.56&4.14&22.00&6.10&44.55&7.88 &21.55&5.74&60.19&11.62\\ 
pre-act. ResNet \cite{he2016identity}&23.77&7.04&25.56&4.14&22.11&6.26&44.55&7.88& 21.41&5.78&60.19&11.62 \\ 
iResNet~\cite{duta2020improved}&22.69&6.46&25.56&4.18&21.36&5.63&44.55&7.92 &20.66&5.43&60.19&11.65 \\
NL-ResNet~\cite{wang2018non}  & 22.91 & 6.42 & 36.72 & 6.18 & 21.40 & 5.83 & 55.71& 9.91 & 21.91 & 6.11 & 71.35& 13.66 \\
SE-ResNet \cite{hu2018squeeze} & 22.74 & 6.37 & 28.07 & 4.15 & 21.31 & 5.79 & 49.29 & 7.90 & 21.38 & 5.80& 66.77 & 11.65 \\
ResNeXt \cite{xie2017aggregated}&  22.44 & 6.25  &  25.03 & 4.30  & 21.03  &  5.66 &  44.18 & 8.07   & 20.98  & 5.48  &  59.95  & 11.84 \\ 
PyConvHGResNet& {\bf 21.52} & {\bf 5.94} & 25.23 & 4.61 & {\bf 20.78} & {\bf 5.57} & 44.63 & 8.42 & {\bf 20.64} & {\bf 5.27} & 60.66 & 12.29 \\
PyConvResNet & {\bf 22.12} & {\bf 6.20}&\underline{24.85}&\underline{3.88} & {\bf 20.99} & {\bf 5.53} & \underline{42.31} & \underline{7.31} & {\bf 20.48} & {\bf 5.27} & \underline{56.64} & \underline{10.72}\\
\hline
\end{tabular}}
\vspace{-0.2in}
\end{table}
\addtolength{\tabcolsep}{+1.9pt}

\addtolength{\tabcolsep}{+5pt}
\begin{table}[t]
\centering
\caption{Validation error rates comparison results of PyConvResNet on ImageNet with different training settings, for network depth 50 and 101  {\tiny($^{\dag}$on the already trained model with 224$\times$224 crop, just perform the test on 320$\times$320)}.}
\label{table:pyconv}
\scalebox{0.8}{
\begin{tabular}{l|ccc|ccc|c}
\hline
\multirow{2}{*}{Network}&\multicolumn{3}{c|}{test crop: 224$\times$224 }&\multicolumn{3}{c|}{test crop: 320$\times$320$^{\dag}$ }&\multirow{2}{*}{params}\\ \cline{2-7}
&top-1&top-5&GFLOPs&top-1&top-5& GFLOPs&\\ 
\hline
\hline
PyConvResNet-50 & 22.12 & 6.20 & 3.88 & 21.10 & 5.55 & 7.91 & 24.85 \\
PyConvResNet-50 + augment & {\bf20.56} & {\bf5.31} & 3.88 & {\bf19.41} & {\bf4.75} & 7.91 & 24.85 \\
\hline
PyConvResNet-101 & 20.99 & 5.53 & 7.31 & 20.03  & 4.82 & 14.92 & 42.31 \\
PyConvResNet-101 + augment & {\bf19.42} & {\bf4.87} & 7.31 & {\bf18.51}  & {\bf4.28} & 14.92 & 42.31 \\
\hline
\end{tabular}}
\end{table}
\addtolength{\tabcolsep}{-5pt}

Fig.~\ref{fig:curves_pyconv} shows  the training and validation curves for comparing our networks, PyConvResNet and PyConvHGResNet,  with baseline ResNet over 50, 101 and 152 layers, where we can notice that our networks significantly improve the learning convergence.  For instance, on 50 depth, on first interval (first 30 epochs, before the first reduction of the learning rate), our PyConvResNet needs less than 10 epochs to outperform the best results of ResNet on all first 30 epochs. Thus, because of improved learning capabilities, our PyConvResNet can require significantly less epochs for training to outperform the baseline.
Table~\ref{table:pyconv_sota} presents the comparison results of our proposed networks with other state-of-the-art networks on 50, 101, 152 layers. Our networks outperform the baseline ResNet~\cite{he2016deep} by a large margin on all depths. For instance, our PyConvResNet improves the top-1 error rate from 23.88 to 22.12 on 50 layers, while having lower number of parameters and FLOPs. Remarkably, our PyConvHGResNet with 50 layers outperforms ResNet with 152 layers on top-1 error rate.  Besides providing better results than pre-activation ResNet~\cite{he2016identity}  and ResNeXt~\cite{xie2017aggregated}, our networks outperform more complex architectures, like SE-ResNet~\cite{hu2018squeeze}, despite that it uses an additional squeeze-and-excitation block, which increases model complexity.

The above results mainly aim to show the advantages of our PyConv over the standard convolution by running all the networks with the same standard training settings for a fair comparison. Note that there are other works which report better results on ImageNet, such as \cite{mahajan2018exploring,tan2019efficientnet,touvron2019fixing}. However, the improvements are mainly due to the training settings. For instance, \cite{tan2019efficientnet}  uses very complex training settings, such as, complex data augmentation (autoAugment \cite{cubuk2019autoaugment}) with different regularization techniques (dropout \cite{srivastava2014dropout}), stochastic depth \cite{huang2016deep}, the training is performed on a powerful Google TPU computational architecture over 350 epochs with a large batch of 2048. The works \cite{mahajan2018exploring,touvron2019fixing}, besides using  a strong computational architecture with many GPUs, take advantage of a large dataset of 3.5B images collected from Instagram (this dataset is not publicly available). Therefore, these resources are not handy to everyone. However, the  results show that PyConv is superior to standard convolution and combining it with~\cite{mahajan2018exploring,tan2019efficientnet,touvron2019fixing} can bring further improvements.  While on ImageNet we do not have access to such scale of computational and data resources to  directly compete with state-of-the-art, we do push further and show that our proposed framework obtains state-of-the-art results on challenging task of image segmentation.

To support our claim, that our networks can be easily improved using more complex training settings, we integrate an additional data augmentation, CutMix~\cite{yun2019cutmix}. As CutMix requires more epochs to converge,  we increase the training epochs to 300 and use a cosine scheduler~\cite{loshchilov2016sgdr} for learning rate decaying. To speed-up the training, we increase the batch size to 1024 and use mixed precision~\cite{micikevicius2017mixed}. Table~\ref{table:pyconv} presents the comparison results of PyConvResNet for the baseline training settings and with the CutMix data augmentation. For both depths, 50- and 101-layers, just adding these simple additional training settings improve significantly the results. For the same trained models, in addition to the standard test crop size of 224$\times$224 we also run  the testing on 320$\times$320 crop size. This results show that there is still room for improvement if more complex training settings are included (as the training settings from \cite{tan2019efficientnet}) and/or additional data used for training (as in \cite{mahajan2018exploring,touvron2019fixing}), however, this requires significantly more computational and data resources, which are not easily available.

\addtolength{\tabcolsep}{+0.5pt}
\begin{table*}[t]
\centering
\caption{Head-to-Head comparison on image segmentation (ResNet-50 as backbone) on {\small ADE20K}.}
\vspace{-0.1in}
\label{table:head-to-head}
\scalebox{0.8}{
\begin{tabular}{l|cccc|cccc}
\hline
\multirow{2}{*}{Head}&\multicolumn{4}{c|}{output stride backbone: 8 }&\multicolumn{4}{c}{output stride backbone: 16}\\
\cline{2-9}
&mean IoU& pixel Acc.& \scalebox{0.8}{params} &\small \scalebox{0.7}{GFLOPs}& mean IoU& pixel Acc.& \scalebox{0.8}{params}& \scalebox{0.7}{GFLOPs}\\ \hline
baseline \cite{zhao2017pyramid}: $3$$\times$$3$ conv& 37.87 & 78.17 & 35.42 &  131.37& 36.84 & 77.84 & 35.42 & 39.52 \\
DeepLabv3 \cite{chen2017rethinking}: ASPP  & 40.91 & 79.92 & 41.48 & 151.17 & 40.34 & 79.44 &  41.48&  44.47  \\
PSPNet \cite{zhao2017pyramid}: PPM & 41.24 & 80.01 & 49.06 & 165.42  & 39.75 & 79.17 &  49.06& 48.08 \\
PyConvSegNet: PyConvPH & {\bf 41.54} & {\bf 80.18} &  \underline{34.40}& \underline{116.84} & {\bf 40.43} & {\bf 79.45} & \underline{34.40}  & \underline{36.08} \\ 
\hline
\end{tabular}}
\vspace{-0.2in}
\end{table*}
\addtolength{\tabcolsep}{-0.5pt}

{\bf PyConv results on semantic segmentation.} We compare our proposed framework, PyConvSegNet, with two of the most powerful  architectures for semantic segmentation \cite{zhao2017pyramid} and \cite{chen2017rethinking}.
Table~\ref{table:head-to-head} presents head-to-head comparison of our method with state-of-the-art heads on image segmentation: PSPNet with Pyramid Pooling Module (PPM) head, and DeepLabv3 with Atrous Spatial Pyramid~Pooling (ASPP).  The baseline is constructed as in~\cite{zhao2017pyramid}, which as head, it basically applies a $3$$\times$$3$ conv over the output feature maps provided by the backbone.
For a fair and direct comparison, all methods use the same auxiliary loss (deep supervision) exactly as in~\cite{zhao2017pyramid}. For a comprehensive view, the reports in terms of number of parameters and FLOPs include the auxiliary loss components. As \cite{zhao2017pyramid} uses an output stride for the backbone of 8 and  \cite{chen2017rethinking} uses 16, we report the experiments for both cases. We run these experiments using the ResNet with 50 layers as backbone. Table~\ref{table:head-to-head} shows that our proposed head is not only more accurate than the other methods, but it is also more efficient, requiring significantly smaller number of parameters and FLOPs than  \cite{zhao2017pyramid} and \cite{chen2017rethinking}. We can also see that without a strong head on top of the backbone, the baseline reports significantly worse results. 

\begin{table}[h]
\vspace{-0.2in}
\noindent\begin{minipage}{.59\textwidth}
\begin{center}
\caption{PyConvSegNet results with different backbones.}
\label{table:seg_depth}
\scalebox{0.7}{
\centering
\begin{tabular}{l|cc|cc|cc}
\hline
\multirow{2}{*}{Backbone}&\multicolumn{2}{c|}{ mean IoU{\small (\%)}} & \multicolumn{2}{c|}{pixel Acc.{\small (\%)}}&\multirow{2}{*}{ \small params}&\multirow{2}{*}{\small GFLOPs}\\
\cline{2-5}
&SS&MS&SS&MS& &  \\
\hline
\hline
ResNet-50&41.54& 42.88&80.18 &80.97 & 34.40 & 116.84 \\ 
PyConvResNet-50&42.08& 43.31&80.31 & 81.18 &  33.69&114.18  \\ 
\hline
ResNet-101&42.88& 44.39&80.75 &81.60 &  53.39& 185.47  \\ 
PyConvResNet-101&42.93& 44.58&80.91 & 81.77& 51.15 & 177.29  \\ 
\hline
ResNet-152&44.04& 45.28&81.18 & 81.89&  69.03& 242.00 \\
PyConvResNet-152&{\bf44.36}& {\bf45.64}&{\bf81.54} & {\bf82.36}&  65.48& 229.11 \\
\hline
\end{tabular}}
\end{center}

\vspace{+0.13in}
Table~\ref{table:seg_depth} shows the results of PyConvSegNet using different depths of the backbones ResNet and PyConvResNet. Besides the single-scale (SS) inference results, we show also the results using multi-scale inference (MS) (scales equal to \{0.5, 0.75, 1, 1.25, 1.5, 1.75\}). 
Table~\ref{table:sota_seg} presents the comparisons of our approach with the state-of-the-art on both validation and testing sets. Notably, our approach PyConvSegNet, with 152 layers for backbone, outperforms PSPNet~\cite{zhao2017pyramid} with its 269-layers heavy backbone, which also requires significantly  more parameters and FLOPs for their PPM head.
\vspace{-0.16in}
\end{minipage}
\addtolength{\tabcolsep}{-4pt}
\begin{minipage}{.4\textwidth}
\vspace{+0.15in}
\centering
\caption{State-of-the-art comparison on ADE20K (single model). {\tiny ($^{\dag}$ increase the crop size just for inference from $473$$\times$$473$ to $617$$\times$$617$;  {\tiny $^{\clubsuit}$} just increase training epochs from 100 to 120 and train over training+validation sets; the results on testing set are provided by the official evaluation server, as the labels are not publicly available. The score is the average of mean IoU and pixel Acc. results.)}}
\label{table:sota_seg}
\scalebox{0.68}{
\centering
\begin{tabular}{l|cc|ccc}
\hline
\multirow{2}{*}{Method}&\multicolumn{2}{c|}{Validation set} & \multicolumn{3}{c}{Testing set}\\
\cline{2-6}
&mIoU& pAcc.& mIoU&pAcc.&Score  \\
\hline
FCN \cite{long2015fully}&29.39  &71.32  &-  &-  &44.80 \\
DilatedNet~\cite{yu2015multi}& 32.31 & 73.55 &  -& - &45.67 \\
SegNet~\cite{badrinarayanan2017segnet}&21.64  & 71.00 & - & - &40.79 \\
RefineNet~\cite{lin2017refinenet}&40.70  & - & - & - & -\\
UperNet~\cite{xiao2018unified}& 41.22 & 79.98 & - & - & -\\
PSANet~\cite{zhao2018psanet}&  43.77& 81.51 & - & - & -\\
KE-GAN~\cite{qi2019ke}& 37.10 & 80.50 & -  & - & -\\
CFNet~\cite{zhang2019co}& 44.89 & - & - & - & -\\
CiSS-Net~\cite{zhou2019context}&42.56  & 80.77 & - & - & -\\
EncNet~\cite{zhang2018context}& 44.65 &81.69  & - &-  &55.67 \\
\hdashline
PSPNet-152 \cite{zhao2017pyramid}&43.51  & 81.38 & - & - &- \\
PSPNet-269 \cite{zhao2017pyramid}&  44.94& 81.69 & - & - & 55.38 \\
\hline
PyConvSegNet-152 &45.64 & 82.36 & 37.75 & 73.61 & 55.68\\
PyConvSegNet-152 $^{\dag}$ &{\bf45.99} & {\bf82.49} & - & - & -\\
PyConvSegNet-152 $^{\clubsuit}$ &- & - & {\bf39.13} & {\bf73.91} & {\bf 56.52}\\
\hline
\end{tabular}}
\end{minipage}
\end{table}
\addtolength{\tabcolsep}{4pt}

\section{Conclusion}
 In this paper we proposed pyramidal convolution (PyConv), which contains several levels of kernels with varying scales. PyConv shows significant improvements for different visual recognition tasks and, at the same time, it is also efficient and flexible, providing a very large pool of potential network architectures. Our novel  framework for image segmentation provides state-of-the-art results.  In addition to a broad range of visual recognition tasks, PyConv can have a significant impact in many other directions, such as  image restoration, completion/inpainting, noise/artifact removal, enhancement and image/video super-resolution. 

\appendix
\section{Appendix\label{sec:appendix}}

In this Appendix we  present additional experiments, architectures details and/or analysis.   It contains three main sections: Section~\ref{sec:obj} presents the details of our architecture for object detection; Section~\ref{sec:vid} presents the details for the video classification pipeline; Finally, Section~\ref{sec:vis_ex} shows some visual examples on image segmentation. 

\subsection{PyConv on object detection\label{sec:obj}}
As we already presented in the the main paper the final result on object detection, that we outperform the baseline by a significant margin (see main contribution (4) in the main paper), this section provides the details of our architecture on object detection and the exact numbers of the results.

\begin{figure*}[t]
  \centering
  \includegraphics[width=1\textwidth]{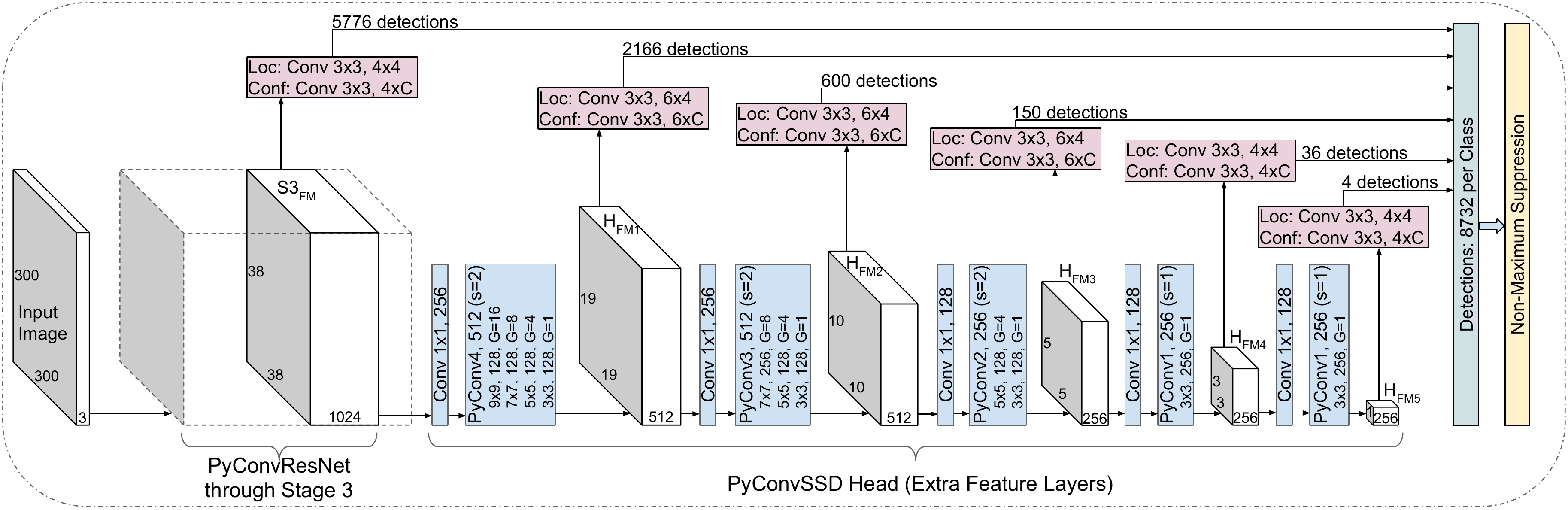}
  \caption{PyConvSSD framework for object detection.}
  \label{fig:pyconvSSD}
\end{figure*}

As our proposed PyConv uses different levels of kernel sizes in parallel, it can provide significant benefits for object detection task, where the objects can appear in the image at different scales.  For object detection, we integrate our PyConv in a powerful approach, Single Shot Detector (SSD)~\cite{liu2016ssd}. SSD is a very efficient single stage framework for object detection, which performs the detection at multiple feature maps resolutions. Our proposed framework for object detection, PyConvSSD,  is illustrated in~Fig.~\ref{fig:pyconvSSD}. The framework contains two main parts:

(1) {\bf PyConvResNet Backbone.} In our framework we use the proposed PyConvResNet as backbone, which was previously pre-trained on ImageNet dataset~\cite{russakovsky2015imagenet}. To maintain a high efficiency of the framework, and also to heave a similar number of output feature maps as in the backbone used in~\cite{liu2016ssd}, we remove from our PyConvResNet backbone all layers after the third stage. We also set all strides in the stage 3 of the backbone network to 1. With this, PyConvResNet  provides (as output of the  stage 3) 1024 output feature maps ($S3_{FM}$) with the spatial resolution $38$$\times$$38$ (for an input image size of $300$$\times$$300$).

(2) {\bf PyConvSSD Head.} Our PyConvSSD head illustrated in~Fig.~\ref{fig:pyconvSSD} uses the proposed PyConv to further extract different features using  different kernel sizes in parallel. Over the resulted  feature maps for the third stage of the backbone we apply a PyConv with four levels (kernel sizes: $9$$\times$$9$, $7$$\times$$7$, $5$$\times$$5$, $3$$\times$$3$). Also PyConv performs the downsampling (stride $s$$=$$2$) of the feature maps using these multiple kernel sizes in parallel. As the feature maps resolution decreases we also decrease the levels of the pyramid for PyConv. The last two PyConv contains only one level (which is basically the standard $3$$\times$$3$) as the spatial resolution of the feature maps is very small. Note that the  last two PyConvs use a stride $s$$=$$1$ and the spatial resolution is decreased just by not using padding. Thus, the head decreases the spatial resolution of the feature maps from $38$$\times$$38$ to $1$$\times$$1$. All the output feature maps from the PyConvs in the head are used for detections.

\addtolength{\tabcolsep}{-8.3pt}
\begin{table*}[t]
\centering
\caption{PyConvSSD with $300$$\times$$300$ input image size (results on COCO val2017).}
\label{table:obj_det}
\scalebox{0.82}{
\begin{tabular}{l|ccc|ccc|ccc|ccc|cc}
\hline
\multirow{2}{*}{Architecture}& \multicolumn{3}{c|}{Avg. Precision, IoU:}& \multicolumn{3}{c|}{Avg. Precision, Area:}& \multicolumn{3}{c|}{Avg. Recall, \#Dets:}& \multicolumn{3}{c|}{Avg. Recall, Area:}&\multirow{2}{*}{params}&\multirow{2}{*}{\scalebox{0.7}{ GFLOPs}} \\
\cline{2-13}
& 0.5:0.95 & 0.5&0.75&S&M&L&1&10&100&S&M&L&& \\
\hline
\hline
Baseline SSD-50& 26.20  &  43.97 &  26.96&   8.12&   28.22&   42.64&  24.50&   35.41&   37.07&   12.61&  40.76&     57.25&  22.89 & 20.92 \\
PyConvSSD-50& \bf29.16  &  \bf47.26 &  \bf30.24&   \bf9.31&   \bf31.21&   \bf47.79&  \bf26.14&   \bf37.81&   \bf39.61&   \bf13.79&  \bf43.87&     \bf60.98&  21.55 & 19.71 \\

\hline
   
Baseline SSD-101&29.58   & 47.69  & 30.80 &  9.38 &   31.96&   47.64&  26.47&   38.00&   39.64&   14.09& 43.54 &     61.03&   41.89& 48.45 \\

PyConvSSD-101 & \bf31.27  & \bf50.00 & \bf32.67 &  \bf10.65 &  \bf33.76 & \bf51.75 & \bf27.33 & \bf39.33  &  \bf41.07 & \bf15.48  & \bf45.53&  \bf63.44   &  39.01 & 45.02 \\
      
\hline
\end{tabular}}
\end{table*}
\addtolength{\tabcolsep}{+8.3pt}

For each of the six output feature maps selected for detection $\{$$S3_{FM},$ $ H_{FM1},$ $H_{FM2},$ $ H_{FM3},$ $ H_{FM4},$ $ H_{FM5}\}$  the framework performs the detection using a coresponding number of default boxes (anchor boxes) $\{$$4,$ $6,$ $6,$ $6,$ $4,$ $4$$\}$ for each spatial location. For instance, for ($S3_{FM}$) output feature maps with the spatial resolution $38$$\times$$38$, using the four default boxes on each location results in 5776 detections. 
For localizing each bounding box, there are four values  that  network should predict (loc: $\Delta(cx, cy, w, h)$, where $cx$ and $cy$ represent the center point of the bounding box, $w$  and $h$ the width and height of the bounding box). This bounding box offset output values are measured relative to a default box position, relative to each feature maps location. Also, for each bounding box, the network should output the confidences for each class category (in total $C$ class categories). For providing the detections the framework uses a classifier which is represented by a $3$$\times$$3$ convolution, that outputs for each bounding box the confidences for all class categories ($C$). For localization the framework uses also a  $3$$\times$$3$ convolution to output the four localization values for each regressed default bounding box. In total, the framework outputs 8732 detections (for $300$$\times$$300$ input image size), which pass through a non-maximum suppression to provide the final detections.

Different from  the original SSD framework~\cite{liu2016ssd}, for a fair and direct comparison, in the baseline SSD,  we replaced the VGG backbone~\cite{simonyan2014very}  with ResNet~\cite{he2016deep}, as ResNet is far superior to VGG in terms of recognition performance and computational costs as shown in \cite{he2016deep}. Therefore, as main differences from our PyConvSSD, the baseline SSD in this work uses ResNet~\cite{he2016deep} as backbone and the SSD head uses standard $3$$\times$$3$ conv (instead of PyConv) as in the original framework~\cite{liu2016ssd}. For showing the exact numbers to compare our PyConvSSD with the baseline on object detection, we use COCO dataset~\cite{lin2014microsoft}, which contains 81 categories. We use for training COCO train2017 (118K images) and for testing COCO val2017 (5K images). We train for 130 epochs using 8 GPUs with 32 batch size each, resulting in 60K training iterations. We use for training SGD optimiser with momentum 0.9, weight decay 0.0005, with the learning rate 0.02 (reduced by 1/10 before 86-th and 108-th epoch). We also use a linear warmup in the first epoch~\cite{goyal2017accurate}. For data augmentation, we perform random crop as in~\cite{liu2016ssd}, color jitter and horizontal flip. We use an input image size of  $300$$\times$$300$ and report the metrics as in~\cite{liu2016ssd}.

Table~\ref{table:obj_det} shows the comparison results of PyConvSSD with the baseline, over 50- and 101-layers backbones.  While being more efficient in terms of number of parameters and FLOPs, the proposed PyConvSSD reports significant improvements over the baseline over all metrics. Notably, PyConvSSD with 50 layers backbone is even competitive with the baseline using 101 layers as backbone. This results show a grasp of the benefits for PyConv on object detection task.

\subsection{PyConv on video classification\label{sec:vid}}
In the main paper we introduced the main result for video classification, that we report significant results over the baseline (see main contribution (4)). This section presents the details of the architecture and the exact numbers. 
PyConv can show significant benefits on video related tasks as it can enlarge the receptive field and process the input using multiple kernels scales in parallel not only spatially but also in the temporal dimension.  Extending the networks from image recognition to video involves extending the 2D spatial convolution to 3D spatio-temporal convolution. Table~\ref{table:net_video} presents the baseline network ResNet3D and our proposed network PyConvResNet3D, which are the initial 2D networks extended to work with video input. The input for the network is represented by 16-frame input clips, with spatial  size  is $224$$\times$$224$. As the temporal size is smaller than spatial dimensions, for our PyConv we do not need to use equally large size on the upper layers of the pyramid. In the first stage of the network, our PyConv with four layers contains kernel sizes of:  7$\times$9$\times$9, 5$\times$7$\times$7, 3$\times$5$\times$5 and 3$\times$3$\times$3 (the temporal dimension comes first).

For video classification, we perform the experiments on Kinetics-400 \cite{kay2017kinetics}, which is a large-scale video recognition dataset that contains  $\sim$246k training videos and 20k validation videos, with 400 action classes. Similar to image recognition, use the SGD optimizer with a standard momentum of 0.9 and weight decay of 0.0001,  we train the model for 90 epochs, starting with a learning rate of 0.1 and reducing it by 1/10 at the 30-th, 60-th and 80-th epochs, similar to \cite{he2016deep,goyal2017accurate}. The models are trained from scratch, using the weights initialization of \cite{he2015delving} for all convolutional layers; for training we use a minibatch of 64 clips over 8 GPUs. Data augmentation is similar to~\cite{simonyan2014very,wang2018non}. For training,  we randomly select 16-frame input clips from the video.  We also skip four frames to cover a longer video period within a clip. The spatial size is 224$\times$224, randomly cropped from a scaled video, where
\begin{multicols}{2}

\renewcommand{\arraystretch}{0.955}
\addtolength{\tabcolsep}{-6pt}
\begin{wraptable}{l}{1\linewidth}
\centering
\caption{Video recognition networks.}
\label{table:net_video}
\scalebox{0.61}{
\centering
\begin{tabular}{c|c|c|c}
\hline
stage                  & output                 & ResNet3D-50& PyConvResNet3D-50 \\
\hline
\multirow{2}{*}{}                 & 16$\times$112$\times$112                & \splitcell{ 5$\times$7$\times$7, 64 \\stride (1,2,2)} & \splitcell{5$\times$7$\times$7, 64 \\ stride (1,2,2)}     \\  
\cline{2-4}
 &  & \splitcell{1$\times$3$\times$3 max pool \\ stride (1,2,2)} & \\  
\hline
1&16$\times$56$\times$56 & \bsplitcell{1$\times$1$\times$1, 64\\ 3$\times$3$\times$3, 64\\ 1$\times$1$\times$1, 256}$\times$3 &
\bsplitcell{1$\times$1$\times$1, 64\\ \underline{PyConv4, 64:}\\  \bsplitcell{7$\times$9$\times$9, 16, G=16\\ 5$\times$7$\times$7, 16, G=8 \\ 3$\times$5$\times$5, 16, G=4 \\ 3$\times$3$\times$3, 16, G=1} \\ 1$\times$1$\times$1, 256}$\times$3 \\ 
\hline

2 & 16$\times$28$\times$28 & \bsplitcell{1$\times$1$\times$1, 128\\ 3$\times$3$\times$3, 128\\ 1$\times$1$\times$1, 512}$\times$4 &
\bsplitcell{1$\times$1$\times$1, 128\\ \underline{PyConv3, 128:}\\  \bsplitcell{5$\times$7$\times$7, 64, G=8 \\ 3$\times$5$\times$5, 32, G=4 \\ 3$\times$3$\times$3, 32, G=1}\\ 1$\times$1$\times$1, 512}$\times$4\\ 
\hline

3 & 8$\times$14$\times$14 & \bsplitcell{1$\times$1$\times$1, 256\\ 3$\times$3$\times$3, 256\\ 1$\times$1$\times$1, 1024}$\times$6  &
\bsplitcell{1$\times$1$\times$1, 256\\ \underline{PyConv2, 256:}\\  \bsplitcell{3$\times$5$\times$5, 128, G=4 \\ 3$\times$3$\times$3, 128, G=1}\\ 1$\times$1$\times$1, 1024}$\times$6 \\ \hline

4 & 4$\times$7$\times$7 & \bsplitcell{1$\times$1$\times$1, 512\\ 3$\times$3$\times$3, 512\\ 1$\times$1$\times$1, 2048}$\times$3 &
\bsplitcell{1$\times$1$\times$1, 512\\ \underline{PyConv1, 512:}\\  \bsplitcell{3$\times$3$\times$3, 512, G=1}\\ 1$\times$1$\times$1, 2048}$\times$3 \\ \hline
 & 1$\times$1$\times$1 &  \begin{tabular}[c]{@{}c@{}}global avg pool\\ 400-d fc\end{tabular}  &  \begin{tabular}[c]{@{}c@{}}global avg pool\\ 400-d fc\end{tabular} \\ \hline

\multicolumn{2}{c|}{\# params}                  &
\bf{47.00}  $\times$ $10^6$                 &
\bf{44.91}  $\times$ $10^6$\\
\hline

\multicolumn{2}{c|}{FLOPs}                      & 
\bf{93.26}  $\times$ $10^9$                   &
\bf{91.81}  $\times$ $10^9$\\
\hline
\end{tabular}}
\end{wraptable}
\addtolength{\tabcolsep}{+6pt}

\addtolength{\tabcolsep}{-9pt}
\begin{wraptable}{r}{1\linewidth}
\vspace{-0.17in}
\centering
\caption{Video recognition error rates {\small(\%)}.}
\label{table:res_video}
\scalebox{0.92}{
\begin{tabular}{l|cccc}
\hline
Architecture &top-1&top-5&params &{\tiny  GFLOPs}\\ \hline
ResNet3D-50 \cite{he2016deep}& 37.01  & 15.41&47.00 & 93.26 \\
PyConvResNet3D-50&   \bf 34.56& \bf13.34&  44.91 & 91.81 \\
\hline
\end{tabular}}
\addtolength{\tabcolsep}{+9pt}

\vspace{0.1in}
  \centering
  \includegraphics[width=0.464\textwidth]{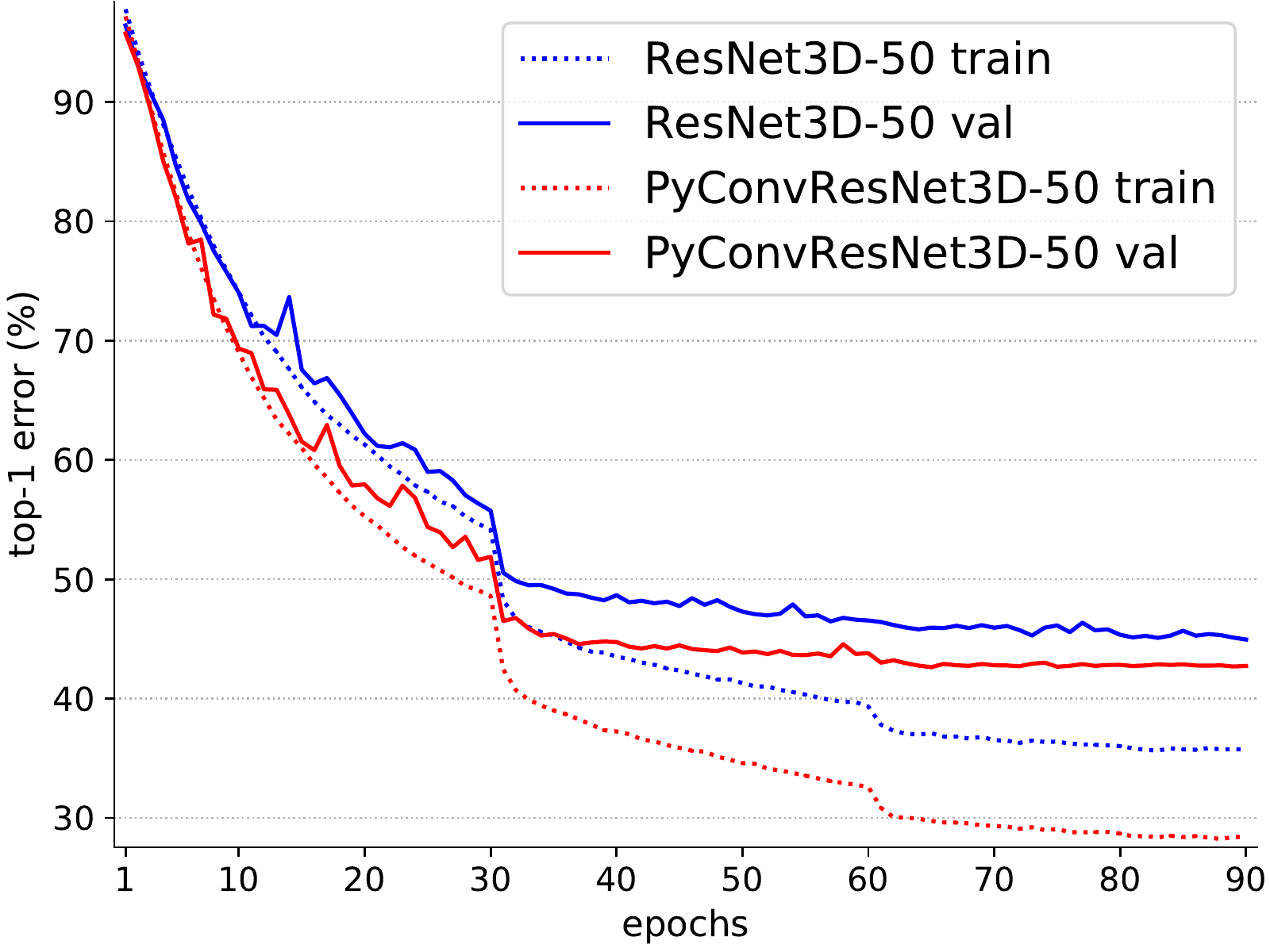}
 \captionof{figure}{Training and validation curves on Kinetics-400 dataset (these  results are computed during training over independent clips).}
  \label{fig:curves_video}
\end{wraptable}
\hfill
\end{multicols}
 the shorter side is randomly selected from the interval [256, 320], similar to \cite{simonyan2014very,wang2018non}.  As the networks on video data are prone to overfitting due to the increase in number of parameters, we use dropout \cite{hinton2012improving} after the global average pooling layer, with a 0.5 dropout ratio. For the final validation, following common practice, we uniformly select a maximum of 10 clips per video. Each clip is scaled  to 256 pixels for the shorter spatial side. We take 3 spatial crops to cover the spatial dimensions. In total, this results in a  maximum  of 30 clips per video, for each of which we obtain a prediction. To get the final prediction for a video, we average the softmax scores. We report both, top-1 and top-5 error rates.

Table~\ref{table:res_video} presents the result comparing our network,  PyConvResNet3D, with the baseline over 50-layers depth. PyConvResNet3D improves significantly the results over baseline, for top-1 error, from 37.01\% to 34.56\%. In the same time our network requires less number of parameters and FLOPs than the baseline.  Fig.~\ref{fig:curves_video} shows the training and validation curves where we can see that our network improves significantly the training convergence. This results show the potential of PyConv on video related tasks.

\subsection{Qualitative examples on image segmentation \label{sec:vis_ex}}
\begin{figure*}[t]
  \centering
  \includegraphics[width=1\textwidth]{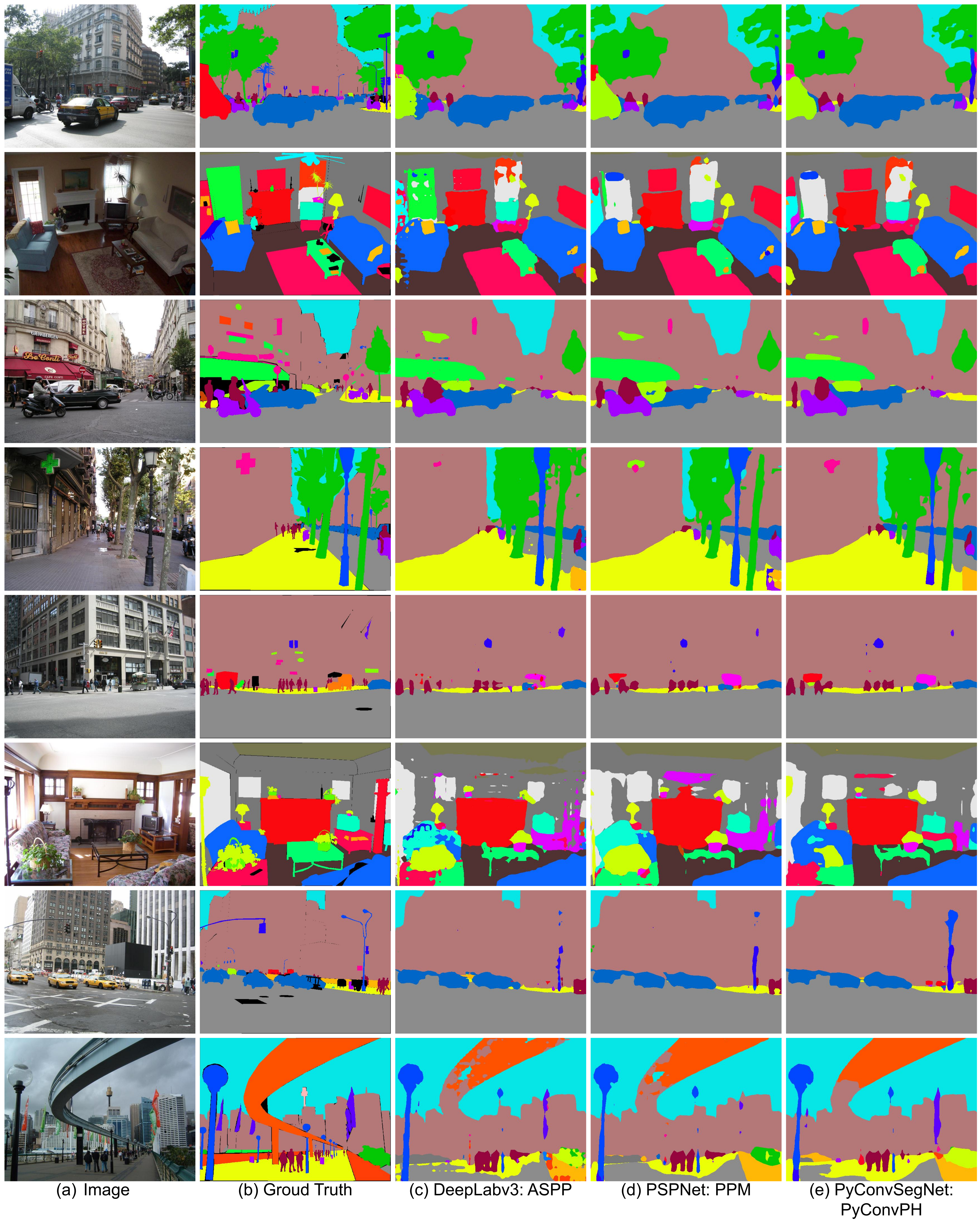}
  \caption{Visual comparison results of our approach PyConSegNet (with PyConvPH head) with state-of-the-art approaches: PSPNet~\cite{zhao2017pyramid} (with PPM head) and  DeepLabv3~\cite{chen2017rethinking} (with ASPP head). The images are from ADE20K dataset~\cite{zhou2019semantic} validation.}
  \label{fig:examples_heads}
\end{figure*}

Fig.~\ref{fig:examples_heads} shows  some qualitative examples for visually comparing our proposed approach for image segmentation, PyConSegNet, with state-of-the-art approaches  PSPNet~\cite{zhao2017pyramid} and DeepLabv3~\cite{chen2017rethinking}. For the numeric results, refer to Table 4 in the main paper (for the output stride backbone 8). 
This examples show the visual comparison results between our proposed head, PyConvPH (PyConv parsing head), with ASPP  (Atrous Spatial Pyramid Pooling) of~\cite{chen2017rethinking} and PPM head (Pyramid Pooling Module) of~\cite{zhao2017pyramid}.

Very suggestive is the last row example of Fig.~\ref{fig:examples_heads}, where we can clearly notice the difference in segmentation details.
It is remarkable that our proposed head can compete at a high level with other state-of-the art approaches for image segmentation while having significantly less requirements in terms of number of parameters and computational complexity. For instance, in comparison with our PyConSegNet, PSPNet~\cite{zhao2017pyramid} requires  over  40\%   more parameters and FLOPs, while DeepLabv3~\cite{chen2017rethinking} requires over 20\%  more parameters and close  to 30\% more FLOPs.

\begin{figure*}[t]
  \centering
  \includegraphics[width=1\textwidth]{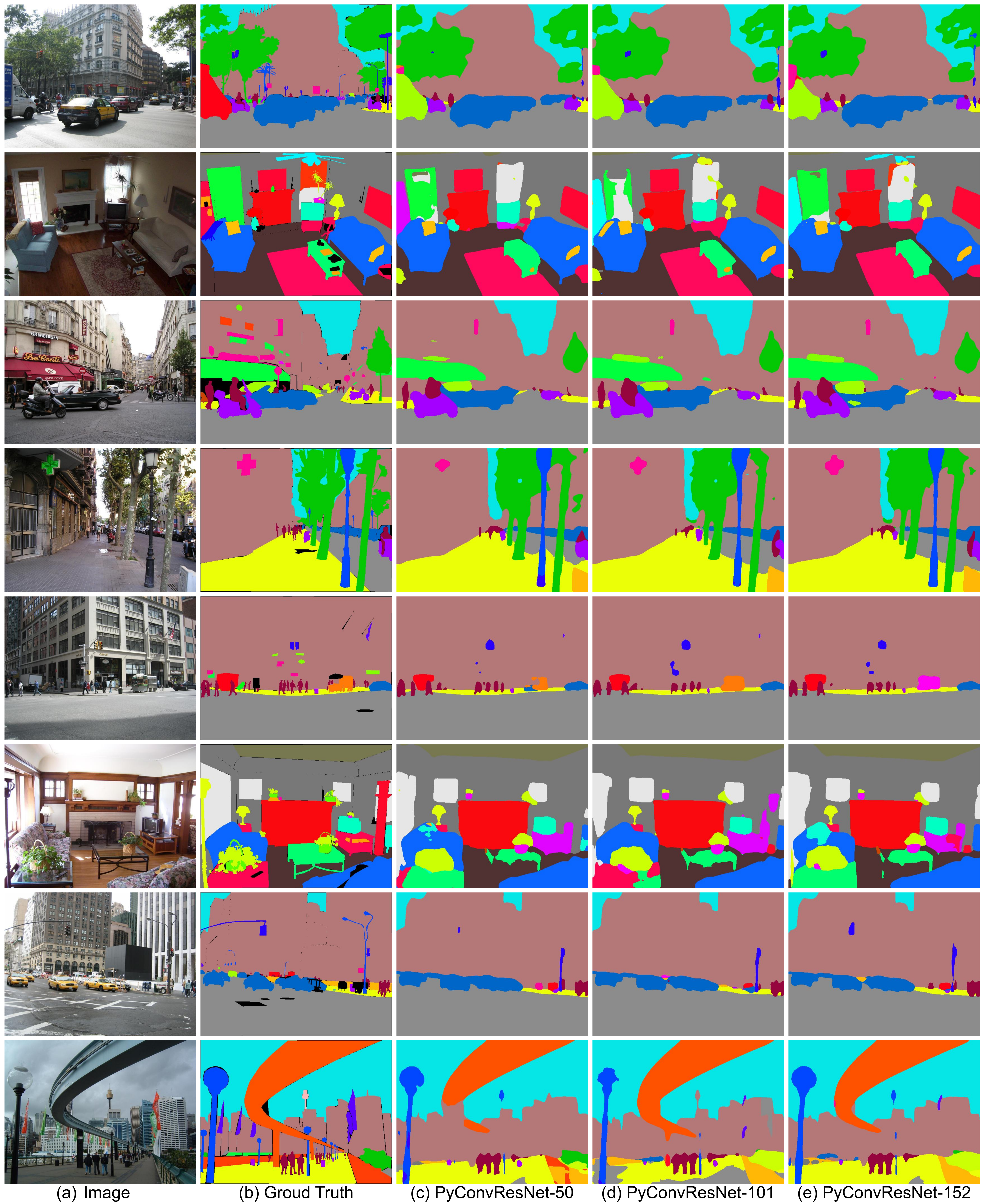}
  \caption{Visual results of our approach, PyConSegNet, on 50-, 101-, 152-layers deep backbone PyConvResNet. The images are from ADE20K dataset~\cite{zhou2019semantic} validation set.  }
  \label{fig:examples_pyconvseg_depths}
\end{figure*}

In the second row example of Fig.~\ref{fig:examples_heads} we can also notice a failure case of our approach, which confuses the door with a  window. However, this case is quite difficult and confusing even for a human eye. Fig.~\ref{fig:examples_pyconvseg_depths} shows some visual results of our approach, PyConSegNet, using 50-, 101-, 152-layers for the PyConvResNet backbone. For the exact number, refer to Table 5 in the main paper (multi-scale inference). Note in the second row of Fig.~\ref{fig:examples_pyconvseg_depths} how the quality of the segmentation for the fan (ceiling mount air fan) is improving while increasing the depth of our PyConvResNet backbone.

\clearpage

\printbibliography

\end{document}